\DocumentMetadata{}
\documentclass[acmsmall,nonacm]{acmart}

\AtBeginDocument{%
  \providecommand\BibTeX{{%
    \normalfont B\kern-0.5em{\scshape i\kern-0.25em b}\kern-0.8em\TeX}}}

\usepackage{tikz}
\usetikzlibrary{positioning, shapes.geometric, arrows.meta, decorations.pathreplacing, calc}
\usepackage{soul}
\usepackage{color}
\usepackage{cancel}

\usepackage{tikz}
\usepackage{svg}
\usepackage{caption}
\usepackage{subfigure}
\usepackage{xspace}
\usepackage{multirow}
\usepackage{stfloats}
\usepackage{algorithm}
\usepackage{algorithmic}
\newcommand{\ours}{{\sc FS-Tree}\xspace}
\newcommand{\ie}{\emph{i.e.,}\xspace}
\newcommand{\aka}{\emph{a.k.a.,}\xspace}
\newcommand{\eg}{\emph{e.g.,}\xspace}

\newcommand{\etc}{\emph{etc.}\xspace}

\setcopyright{acmcopyright}
\copyrightyear{2018}
\acmYear{2018}
\acmDOI{XXXXXXX.XXXXXXX}

\acmJournal{JACM}
\acmVolume{37}
\acmNumber{4}
\acmArticle{111}
\acmMonth{8}

\begin{document}

\setcitestyle{numbers,compress}

\title{Tree-based Models for Vertical Federated Learning: A Survey}

\author{Bingchen Qian}
\email{qianbingchen.qbc@alibaba-inc.com}
\authornote{Co-first authors, listed in alphabetical order.}
\affiliation{%
  \institution{Alibaba Group}
  \country{China}
}

\author{Yuexiang Xie}
\email{yuexiang.xyx@alibaba-inc.com}
\authornotemark[1]
\affiliation{%
  \institution{Alibaba Group}
  \country{China}
}

\author{Yaliang Li}
\authornote{Corresponding author.}
\email{yaliang.li@alibaba-inc.com}
\affiliation{%
  \institution{Alibaba Group}
  \country{USA}
}

\author{Bolin Ding}
\email{bolin.ding@alibaba-inc.com}
\affiliation{%
  \institution{Alibaba Group}
  \country{USA}
}

\author{Jingren Zhou}
\email{jingren.zhou@alibaba-inc.com}
\affiliation{%
  \institution{Alibaba Group}
  \country{USA}
}

\renewcommand{\shortauthors}{Bingchen Qian, Yuexiang Xie, Yaliang Li, Bolin Ding, and Jingren Zhou}

\begin{abstract}
Tree-based models have achieved great success in a wide range of real-world applications due to their effectiveness, robustness, and interpretability, which inspired people to apply them in vertical federated learning (VFL) scenarios in recent years. In this paper, we conduct a comprehensive study to give an overall picture of applying tree-based models in VFL, from the perspective of their communication and computation protocols. We categorize tree-based models in VFL into two types, \ie feature-gathering models and label-scattering models, and provide a detailed discussion regarding their characteristics, advantages, privacy protection mechanisms, and applications. This study also focuses on the implementation of tree-based models in VFL, summarizing several design principles for better satisfying various requirements from both academic research and industrial deployment. We conduct a series of experiments to provide empirical observations on the differences and advances of different types of tree-based models.
\end{abstract}





\maketitle

\section{Introduction}
\label{sec:intro}

Although machine learning models have made remarkable progress over the past few years~\cite{hinton2006reducing, lecun2015deep, chen2016xgboost, he2016identity, krizhevsky2017imagenet, srivastava2014dropout, vaswani2017attention}, driven by large-scale data and powerful computing capabilities~\cite{machine2015jordan, survey2021li}, there exist several factors that hinder the broader application of machine learning models in real-world scenarios: data privacy and decentralization.
On the one hand, the public is becoming increasingly cautious about privacy leakage issues related to the usage of personal data, while more and more completed regulations (\eg GDPR\footnote{https://gdpr-info.eu/}, CCPA\footnote{https://oag.ca.gov/privacy/ccpa}, FISMA\footnote{https://www.cisa.gov/federal-information-security-modernization-act}, \etc) are promulgating to promote that data sharing among different organizations/departments is kept in a privacy-preserving manner.
On the other hand, data required for a real-world application are often inevitably scattered across multiple data owners. Such decentralization scenarios make centralized storage and usage nearly impossible, due to the unaffordable costs of data collecting and issues of multi-party authorities for data accessing.

As a result, how to cooperate across multiple parties (\ie data owners) to train machine learning models while protecting data privacy and keeping data decentralization has attracted rapidly growing popularity in both academic and industrial, and has motivated the proposal of {\it Federated Learning} (FL)~\cite{mcmahan2017communication, kairouz2021advances}.
Concretely, participants involved in FL are suggested to locally train machine learning models based on their private data, and then exchange the learned knowledge (\eg updated models, gradients, \etc) with each other to produce models that are more effective and robust than isolated-trained models.
According to the forms of data partition, FL can be roughly categorized into {\it Horizontal Federated Learning} (HFL) and {\it Vertical Federated Learning} (VFL)~\cite{yang2019federated}. 
HFL refers to the setting where parties share the same feature space but the sets of data samples are almost non-intersecting, while VFL refers to the setting where parties' data samples are overlapped but their feature spaces are different and complementary.

\begin{figure}
    \centering
    \includegraphics[width=0.88\textwidth]{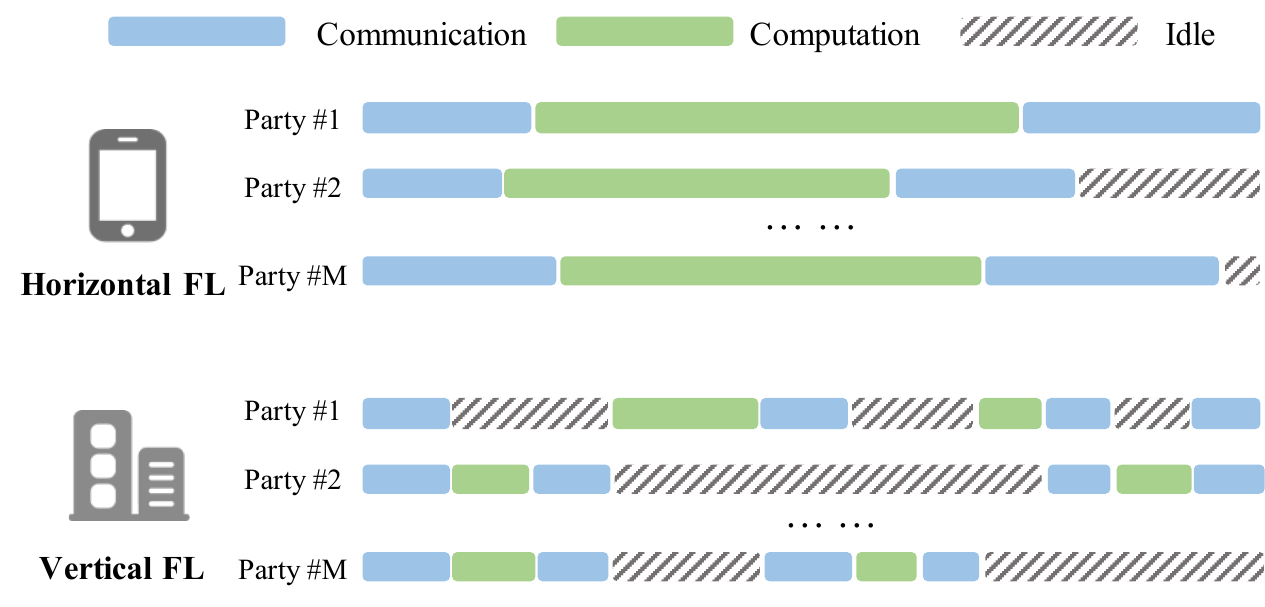}
    \caption{Communication and computation behaviors happen in a training round of Horizontal and Vertical FL.}
    \label{fig:comm_comp}
\end{figure}
The applications of VFL include financial risk management~\cite{chen2021homomorphic, long2020federated, wang2019interpret, cheng2020federated}, joint marketing~\cite{ammad2019federated, zhang2021vertical, cui2021exploiting, shmueli2017secure, leo2019machine}, smart city~\cite{zheng2022applications, jiang2020federated, ramu2022federated}, and so on~\cite{chen2020vafl, teimoori2022secure, chen2020vertically, he2020group, gupta2018distributed}. 
Compared to the HFL scenario where multiple parties train their local models independently, in the VFL scenario, the training process of the entire model is separated into different sub-tasks according to the feature spaces of parties, and thus a party might need to wait for the intermediate results provided by other parties. 
As illustrated in Figure~\ref{fig:comm_comp}, such characteristics of the VFL scenario lead to a phenomenon that a party's computation behaviors (\eg forward propagating and residuals calculating) and communication behaviors (\eg results sending and receiving) are naturally fragmented and alternately executed in each training round.

Various types of methods and techniques have been proposed for the VFL scenario, including kernel-based models~\cite{dang2020large, gu2020federated}, linear models~\cite{kairouz2021advances, yang2019parallel}, deep neural networks~\cite{bonawitz2017practical, romanini2021pyvertical, gupta2018distributed}, and tree-based models~\cite{liu2020federated, cheng2021secureboost, fang2021large, tian2020federboost, li2022opboost, chen2021fed, song2021federated, jin2022towards, zheng2023privet, lu2023squirrel, jiang2024sigbdt, akhavan2023level, xu2024elxgb, chen2022privdt, xia2022privacy}.
Among these methods, tree-based models are particular and essential due to the following reasons. 
Firstly, tree-based models have been proven to be effective and robust on tabular data that might consist of both numerical and categorical features~\cite{grinsztajn2022tree}, which allows tree-based models to be feasible solutions in scenarios where tabular data is widely available, \eg cross-silo VFL. 
Secondly, the interpretability of tree-based models satisfies the demand in many industries that need to ensure the model predictions are extremely reliable, including healthcare, education, finance, and so on. 
Moreover, compared to deep neural networks, tree-based models usually need fewer computation resources and hardware support to achieve competitive performance.
Last but not least, the architectures of tree-based models are compatible with rich types of privacy protection mechanisms, \eg differential privacy~\cite{dwork2008differential, abadi2016deep, dwork2014algorithmic}, homomorphic encryption~\cite{paillier1999public, yi2014homomorphic, fontaine2007survey, acar2018survey, naehrig2011can, gentry2009fully}, and secure multi-party computation~\cite{yao1982protocols, goldreich1998secure, du2001secure, zhao2019secure, evans2018pragmatic, ben2008fairplaymp}, which further broadens the usage of tree-based models in the VFL scenario.

Compared to traditional VFL, tree-based models in VFL exhibit significant differences, including the nature of the heterogeneous messages that must be exchanged among participants (including both feature-related and label-related information), the unique behaviors involved in generating and handling these heterogeneous messages, and rich and robust privacy protection algorithms that need to be implemented to safeguard the sensitive information during these exchanges. 
As a result, tree-based models in VFL necessitate unique designs and classifications that are different from those typically employed in traditional VFL.

These years, researchers have conducted surveys on federated learning~\cite{yang2019federated, li2020federated, kairouz2021advances, aledhari2020federated, li2021survey, liu2022vertical, yu2024survey, yang2023survey, ye2024vertical} to draw an overall picture of FL from the perspectives of definitions, challenges, architectures, approaches, applications, and future directions.
Recently studies~\cite{chatel2021sok, ong2022tree} review the development of tree-based models in FL, and highlight the necessity and advantages of tree-based models compared to other models such as neural networks. 
However, to the best of our knowledge, there is currently no dedicated survey focusing on tree-based models for VFL, which motivates us to provide a comprehensive and up-to-date survey that highlights this promising research direction and aims to further inspire the community.

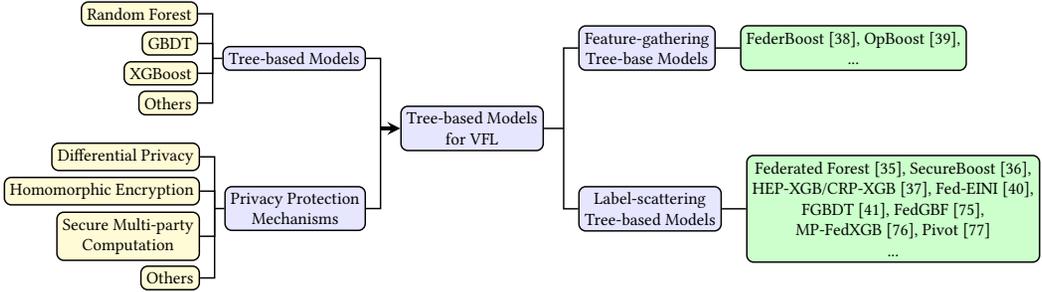
\begin{figure}[t!]
    \centering  
    \vspace{0.5cm}
    \resizebox{\textwidth}{!}{  
    \begin{tikzpicture}[
        node distance=1cm,
        every node/.style={draw, fill=blue!10, rounded corners, align=center},
        left_node/.style={fill=yellow!20},  
        right_node/.style={fill=green!20},  
        >=stealth, 
        line width=0.3mm  
    ]

    \node (tbm_vfl) {Tree-based Models \\ for VFL};

    \node [above left=of tbm_vfl] (tbm) {Tree-based Models};
    \node [below left=of tbm_vfl] (ppm) {Privacy Protection \\ Mechanisms};

    \coordinate (helpertbm) at ($(tbm.east) + (0.3,0)$);
    \coordinate (helperppm) at ($(ppm.east) + (0.3,0)$);
    \coordinate (leftmidPoint) at ($(helpertbm)!0.5!(helperppm)$);  
    \coordinate (lefthelperVert) at (leftmidPoint |- tbm_vfl.west);     

    \draw[-] (tbm.east) -- (helpertbm) -- (lefthelperVert);
    \draw[-] (ppm.east) -- (helperppm) -- (lefthelperVert);
    \draw[->, line width=0.8mm] (lefthelperVert) -- (tbm_vfl.west);  

    \node [above right=of tbm_vfl] (fg) {Feature-gathering \\ Tree-base Models};
    \node [below right=of tbm_vfl] (ls) {Label-scattering \\ Tree-based Models};

    \coordinate (helperfg) at ($(fg.west) + (-0.3,0)$);
    \coordinate (helperls) at ($(ls.west) + (-0.3,0)$);
    \coordinate (rightmidPoint) at ($(helperfg)!0.5!(helperls)$);  
    \coordinate (righthelperVert) at (rightmidPoint |- tbm_vfl.east);     

    \draw[-] (fg.west) -- (helperfg) -- (righthelperVert);
    \draw[-] (ls.west) -- (helperls) -- (righthelperVert);
    \draw[-] (righthelperVert) -- (tbm_vfl.east);

    \node [above left=0.4cm and 0.5cm of tbm, left_node] (rf) {Random Forest};
    \node [above left=-0.2cm and 0.5cm of tbm, left_node] (gbdt) {GBDT};
    \node [below left=-0.2cm and 0.5cm of tbm, left_node] (xgb) {XGBoost};
    \node [below left=0.4cm and 0.5cm of tbm, left_node] (others) {Others};

    \coordinate (helperrf) at ($(rf.east) + (0.3,0)$);
    \coordinate (helpergbdt) at ($(gbdt.east) + (0.3,0)$);
    \coordinate (helperxgb) at ($(xgb.east) + (0.3,0)$);
    \coordinate (helperothers) at ($(others.east) + (0.3,0)$);

    \coordinate (leftmidPointtbm) at ($(helpergbdt)!0.5!(helperxgb)$);  
    \coordinate (lefthelperVerttbm) at (leftmidPointtbm |- tbm.west);   

    \draw[-] (rf.east) -- (helperrf) -- (lefthelperVerttbm);
    \draw[-] (gbdt.east) -- (helpergbdt) -- (lefthelperVerttbm);
    \draw[-] (xgb.east) -- (helperxgb) -- (lefthelperVerttbm);
    \draw[-] (others.east) -- (helperothers) -- (lefthelperVerttbm);

    \draw[-] (lefthelperVerttbm) -- (tbm.west);

    \node [above left=0.3cm and 0.5cm of ppm, left_node] (dp) {Differential Privacy};
    \node [above left=-0.4cm and 0.5cm of ppm, left_node] (he) {Homomorphic Encryption};
    \node [below left=-0.4cm and 0.5cm of ppm, left_node] (smpc) {Secure Multi-party \\ Computation};
    \node [below left=0.7cm and 0.5cm of ppm, left_node] (others_ppm) {Others};

    \coordinate (helperdp) at ($(dp.east) + (0.3,0)$);
    \coordinate (helperhe) at ($(he.east) + (0.3,0)$);
    \coordinate (helpersmpc) at ($(smpc.east) + (0.3,0)$);
    \coordinate (helperothers_ppm) at ($(others_ppm.east) + (0.3,0)$);

    \coordinate (leftmidPointppm) at ($(helperhe)!0.5!(helpersmpc)$);  
    \coordinate (lefthelperVertppm) at (leftmidPointppm |- ppm.west);   

    \draw[-] (dp.east) -- (helperdp) -- (lefthelperVertppm);
    \draw[-] (he.east) -- (helperhe) -- (lefthelperVertppm);
    \draw[-] (smpc.east) -- (helpersmpc) -- (lefthelperVertppm);
    \draw[-] (others_ppm.east) -- (helperothers_ppm) -- (lefthelperVertppm);

    \draw[-] (lefthelperVertppm) -- (ppm.west);

    \node [right=0.5cm of fg, right_node] (ls1) {FederBoost~\cite{tian2020federboost}, OpBoost~\cite{li2022opboost},\\ ...};
    \node [right=0.5cm of ls, right_node] (ls2) {
    Federated Forest~\cite{liu2020federated}, SecureBoost~\cite{cheng2021secureboost}, \\  HEP-XGB/CRP-XGB~\cite{fang2021large}, Fed-EINI~\cite{chen2021fed}, \\
    FGBDT~\cite{song2021federated}, FedGBF~\cite{han2022fedgbf}, \\ MP-FedXGB~\cite{xie2022efficient}, Pivot~\cite{wu2020privacy} \\ ...};

    \draw[-] (ls1.west) -- (fg.east);
    \draw[-] (ls2.west) -- (ls.east);

    \end{tikzpicture}
    }
    \caption{Overview of tree-based models for vertical federated learning.}
    \label{fig:tbm_vfl}
\end{figure}

In this study, as illustrated in Figure~\ref{fig:tbm_vfl}, we give an overview of tree-based models (TBMs) in vertical federate learning (VFL), focusing on the communication and computation protocols, and the privacy protection mechanisms for the information that is required to be shared during the training and inference procedural.
To be more specific, TBMs in VFL can be divided into {\it feature-gathering TBMs} and {\it label-scattering TBMs} according to their communication and computation protocols. 
In order to determine the splitting rules at the nodes of decision trees, feature-gathering TBMs propose that the feature-related information (\eg the ordinal numbers) is sent from the feature owners to the parties who hold labels for calculating the maximum splitting gain, while the label-scattering TBMs suggest the label-related information (\eg the first-order and second-order gradients) is broadcast from the label owners to other parties.
Furthermore, different privacy protection mechanisms are preferred in feature-gathering TBMs and label-scattering TBMs according to the types of shared messages, causing the differences in their performance from the perspectives of communication and computation cost, and protection strength.
Please refer to Section~\ref{sec:method} for more details on the characteristics, differences, and pros and cons of feature-gathering and label-scattering TBMs.

Moreover, this study is also concerned on the implementations of TBMs in the VFL scenario, for both applying the existing works and developing new algorithms. Previous studies~\cite{li2021survey, liu2022unifed} have pointed out that several open-source FL platforms can allow users to apply TBMs in the VFL scenario, such as FATE~\cite{liu2021fate}, Fedlearner, FedTree~\cite{fedtree},  SecretFlow, FederatedScope~\cite{xie2022federatedscope}, and so on.
Taking a step forward, we provide discussions on the key challenges of developing TBMs, and summarize several design principles for making FL platforms more comprehensive and extendable.
We conduct a series of experiments to show the trade-off among model utility, protection effect, and resource cost when applying TBMs in the VFL scenario, which can be a reference for users to choose suitable types of TBMs according to their applications.

\vspace{0.1in}
\noindent {\bf Contributions}.
The main contributions can be summarized as follows:
\begin{itemize}
    \item We propose to categorize the TBMs in the VFL scenario according to the communication and computation protocols, which results in feature-gathering TBMs and label-scattering TBMs. 
    For these two types of TBMs, we provide a detailed description of their training and inference procedure to highlight their differences and advantages.
    \item Based on different computation and communication protocols, we discuss various privacy protection mechanisms that are designed to protect different types of shared information when applying TBMs, and further take a close look at some advanced algorithms.
    \item Towards a better implementation, we summarize the open-source FL platforms that allow users to apply TBMs in the VFL scenario, and point out several design principles to inspire the community for both academic and industrial.
    \item Last but not least, we conduct a series of experiments on widely-used datasets to provide an empirical understanding of the characteristics of tree-based models.
\end{itemize}

\noindent {\bf Paper Organization}.
The rest of this paper is organized as follows.
In Section~\ref{sec:preliminaries}, we provide some preliminaries, including the concepts of VFL, TBMs, and privacy protection. 
Then in Section~\ref{sec:method}, we introduce the details of two different types of TBMs in the VFL scenario, \ie feature-gathering TBMs and label-scattering TBMs, providing the discussions on communication and computation protocols, privacy protection mechanisms, and representative algorithms.
After that, we review the open-source FL platforms and summarize several design principles for supporting TBMs in the VFL scenario, as described in Section~\ref{sec:infrastructure}.
In Section~\ref{sec:exp}, We conduct experiments to provide empirical observations and insights.
Finally, we present real-world applications and highlight future directions in Section~\ref{Sec:applications}, and provide conclusions in Section~\ref{Sec:conclusions}.

\section{Backgrounds}
\label{sec:preliminaries}

\subsection{Vertical Federated Learning}
Vertical federated learning (VFL) involves multiple parties in training machine learning models based on decentralized data collaboratively, where different parties have different feature spaces but their sample spaces are aligned.
These parties can be divided into two categories according to whether they own labels or not. The one party who locally keeps the labels is called {\it task party} (\aka task client) while others are called  {\it data parties} (\aka data clients). 
In some cases, there also exists one or more {\it coordinators} (\aka servers) who play as the trusted third parties. 
Different from the server in the Horizontal Federated Learning (HFL) scenario, the coordinators in the VFL scenario are mostly focused on distributing the necessary information (such as public keys for encryption algorithms) and hardly involve the training process.

Formally, assume that $M$ parties involve in a VFL course,  the $m$-th party locally keeps the dataset $\mathcal{D}_m = \{(x_i^{(m)}, y_i) \in \mathcal{X}_m \times \mathcal{Y} | i \in \mathcal{I}\}$ for a task party and  $\mathcal{D}_m = \{x_i^{(m)} \in \mathcal{X}_m | i \in \mathcal{I}\}$ for a data party, where $ \mathcal{X}_m$ denotes the feature space of the $m$-th party, $\mathcal{Y}$ denotes the label space, and $\mathcal{I}$ denotes the sample space. 
Note that all the participating parties adopt the same sample space $\mathcal{I}$ through aligning techniques like private set intersection~\cite{pinkas2014faster, pinkas2018scalable, dong2013private, chen2017fast}, and their feature spaces are always different and complementary, \ie $\mathcal{X}_{m_1} \neq \mathcal{X}_{m_2}, \forall m_1,m_2 \in [M]$ and $\mathcal{X} = \bigcup_{m=1}^{M} \mathcal{X}_m$. 
Without sharing private data directly,  these $M$ parties aim to train a model $f_\theta: \mathcal{X}\rightarrow \mathcal{Y}$ parameterized by $\theta$, with the loss $l: \mathcal{Y}\times \mathcal{Y}\rightarrow \mathbb{R}^+\cup \{0\}$. The {\it loss function} in VFL can be given as: 
\begin{equation}
    \mathcal{L}=\frac{1}{|\mathcal{I}|}\sum_{i\in \mathcal{I}, (x_i, y_i)\sim \mathcal{X}\times \mathcal{Y}} l(f_\theta(x_i),y_i).
\end{equation}
During the training process, the model parameters $\theta$ (such as the splitting rules in tree-based models) are stored and optimized in multiple parties.

\subsection{Tree-based Models}
Tree-based models (TBMs) in the VFL scenario are mainly constructed with multiple decision trees~\cite{quinlan1986induction, quinlan2014c4, breiman2017classification}, which have been successfully used in a large range of real-world applications for solving classification and regression tasks~\cite{maimon2014data, song2015decision, safavian1991survey, rokach2005top, sharma2016survey}. 

For a decision tree, an internal node (including the root node) represents a splitting rule consisting of a splitting feature and a splitting value, and the branches attached to this internal node indicate the results according to the splitting rule. For example, for a binary decision tree, given an internal node with splitting feature $x$ and splitting value $a$, the branches represent instances with $x\leq a$ and $x > a$ (for numerical features) or $x = a$ and $x \neq a$ (for categorical features), respectively.
Besides the internal nodes, a decision tree includes several leaf nodes to represent the prediction results.
The input data are fed into the root of a decision tree,  partitioned into different subsets according to the splitting rules of the traversing nodes, and finally reach the leaf nodes to derive corresponding predictions.

Several tree-based models used in the VFL scenarios are introduced in the rest of this section, including {\it Random Forest}, {\it Gradient-Boosted Decision Trees} (GBDT), and {\it eXtreme Gradient Boosting} (XGBoost).

\subsubsection{Random forest}
Random forest \cite{breiman2001random} is a bagging-based ensemble supervised machine learning technique, which usually uses {\it Classification and Regression Trees} (CARTs) as weak learners. The randomization of a random forest can be presented in two different ways: random selection of samples and random selection of features. 
After building several CARTs, the predicted results are derived by either majority voting (for classification tasks) or averaging operation (for regression tasks).

\subsubsection{GBDT}
Given a dataset $\mathcal{D} \in \mathbb{R}^{m\times d}$ with $m$ samples and $d$ features, GBDT \cite{friedman2001greedy} predicts the output by using T regression trees, which can be formulated as:
\begin{equation}
 \hat{y}_i=\sum_{t=1}^{T}f_{t}(x_i), ~~f_{t}\in \mathcal{F},
\end{equation}
where $\mathcal{F}=\{f(x)=w_{q(x)}\}$ is a collection of CARTs.
The function $q(x)$ maps each input feature $x$ to the corresponding leaf index, and $w\in\mathbb{R}^{K}$ is the weight vector, where $K$ is the number of leaves in the tree. 
GBDT tries to minimize the following regularized loss function \cite{chen2016xgboost}:
\begin{equation}
    \mathcal{L}\simeq \sum_il(y_i,\hat{y}_i)+\sum_{t}\Omega(f_{t}),
\end{equation}
where $\Omega(f_{t})=\frac{1}{2}\lambda\|w\|^2$ is a regularization term, and $\lambda$ is a hyperparameter to control the strength of regularization.

In order to optimize the above loss function, GBDT minimizes the following function at the $t$-th iteration \cite{si2017gradient}:
\begin{equation}
    \mathcal{L}^{(t)}\simeq \sum_{i=1}^m\left[l(y_i,\hat{y}_i^{(t-1)})+g_if_t(x_i)+\frac{1}{2}f_t^2(x_i)\right]+\Omega(f_t), 
\end{equation}
where $\mathcal{L}^{(t)}$ denotes the $t$-th loss of the training process, $l(y_i,\hat{y}_i^{(t-1)})$ is the loss function of the $i$-th sample between the prediction of the $(t-1)$-th iteration $\hat{y}_i^{(t-1)}$ and the target value $y_i,$ $g_i=\partial_{\hat{y}^{(t-1)}}l(y_i,\hat{y}_i^{(t-1)})$ is the first order gradient statistics on the loss function.

Finally, the optimal weight of the $j$-th leaf node can be given as:
\begin{equation}
    w_j^*=-\frac{\sum_{i\in I_j}g_i}{|I_j|+\lambda},
\end{equation}
where $I_j$ denotes the sample indices that belong to the $j$-th leaf node. 
To find the best splitting threshold for the internal node, GBDT greedily maximizes the following gain score:
\begin{equation}
    Gain=\frac{(\sum_{i\in I_L}g_i)^2}{|I_L|+\lambda}+\frac{(\sum_{i\in I_R}g_i)^2}{|I_R|+\lambda}, \label{eq:gbdt_gain}
\end{equation}
where $I_L$ and $I_R$ are the instance sets of left and right nodes after the split.

\subsubsection{XGBoost}
XGBoost \cite{chen2016xgboost} is an efficient implementation of GBDT.
One of the most important improvements made by XGBoost is that a second-order Taylor expansion function is used to approximate the loss function, as defined by:
\begin{equation}
    \mathcal{L}^{(t)}\simeq \sum_{i=1}^m\left[l(y_i,\hat{y}_i^{(t-1)})+g_if_t(x_i)+\frac{1}{2}h_if_t^2(x_i)\right]+\Omega(f_t), 
\end{equation}
where $h_i=\partial^2_{\hat{y}^{(t-1)}}l(y_i,\hat{y}_i^{(t-1)})$ is the second order gradient statistics on the loss function and $\Omega(f_t)=\gamma T+\frac{1}{2}\lambda\|w\|^2$. Here, $\gamma$ and $\lambda$ represent the regularizers to adjust the number and weights of leaves, respectively.

As a result, the optimal weight of the $j$-th leaf node can be calculated as:
\begin{equation}
    w_j^* = -\frac{\sum_{i\in I_j}g_i}{\sum_{i\in I_j}h_i+\lambda}, \label{eq:xgb_set_weight}
\end{equation}
where $I_j$ denotes the sample indices that belong to the $j$-th leaf node.
In order to find the best splitting threshold for the internal nodes, XGBoost greedily maximizes the following gain score: 
\begin{equation}
    Gain=\frac{1}{2}\left[\frac{(\sum_{i\in I_L}g_i)^2}{\sum_{i\in I_L}h_i+\lambda}+\frac{(\sum_{i\in I_R}g_i)^2}{\sum_{i\in I_R}g_i+\lambda}-\frac{(\sum_{i\in I}g_i)^2}{\sum_{i\in I}h_i+\lambda}\right]-\gamma,
    \label{eq:xgb_gain}
\end{equation}
where $I_L$ and $I_R$ are the instances of left and right nodes after the splitting operation with $I=I_L\cup I_R$.

\subsection{Privacy Protection}
In the VFL scenario, the task party and data parties need to exchange some intermediate results during the process of building decision trees, \eg the feature-related information and the label-related information, which might cause privacy leakage without protection~\cite{tian2020federboost, cheng2021secureboost}.
Therefore, privacy protection techniques are necessary to consider in VFL algorithms and applications, including {\it Differential Privacy} (DP), {\it Homomorphic Encryption} (HE), {\it Secure Multi-Party Computation} (SMPC), and {\it Trusted Execution Environments} (TEE).

\subsubsection{Differential privacy}
Differential privacy \cite{dwork2008differential, abadi2016deep, dwork2014algorithmic} is the technology that enables researchers to avail a facility in obtaining useful information from the databases, containing people's personal information, without divulging the personal identification of individuals. Local DP (LDP) \cite{kasiviswanathan2011can, yang2020local, arachchige2019local, cormode2018privacy} is one type of DP, which is based on clients adding DPs to the data themselves.
The distance-based LDP~\cite{alvim2018local, chatzikokolakis2013broadening, he2014blowfish} is a special kind of LDP, which measures the level of privacy assurance between any pair of sensitive data based on their distance from each other.

\subsubsection{Homomorphic encryption}
Homomorphic encryption \cite{paillier1999public, yi2014homomorphic, fontaine2007survey, acar2018survey, naehrig2011can, gentry2009fully} is a special encryption method that allows the ciphertext to be processed including addition and multiplication to obtain a result that is still encrypted. However, it usually requires more computation resources or storage costs. 
A partially homomorphic encryption (PHE) scheme is a probabilistic asymmetric encryption scheme for restricted computation over the ciphertexts.

\subsubsection{Secure multi-party computation}
Secure Multi-Party Computation~\cite{yao1982protocols, goldreich1998secure, du2001secure, zhao2019secure, evans2018pragmatic, ben2008fairplaymp} is proposed to solve the problem of how to safely compute a conventional function in the absence of a trusted third party. Among the different protocols in SMPC, one of the widely-used techniques is {\it Secret Sharing}~\cite{mohassel2018aby3}. Secret sharing suggests partitioning a secret value into several frames and sending one of the frames to a party, which satisfies that only a certain number of parties can jointly complete the decryption process for exactly recovering the secret value. 
From the perspectives of types, different secret-sharing techniques might allow secret addition, secret multiplication, secret division, and other mathematical operations.

\subsubsection{Trusted execution environments}
Trusted Execution Environments~\cite{garfinkel2003terra, sabt2015trusted} is a separate processing environment with computing and storage capabilities that provide security and integrity protection. The basic idea is that a separate isolated memory is allocated in hardware for sensitive data, all calculations of sensitive data are performed in this memory, and no other part of the hardware can access the information in this isolated memory except for authorized interfaces. In this way, private computation of sensitive data can be achieved.

\section{Tree-based Models for VFL}
\label{sec:method}

Different from the backward propagation of training deep neural networks, the basic optimization step of tree-based models is to find a splitting rule that can achieve the best gain at each internal node, which can be formulated as a function $r=h(x,y)$ where $(x,y)\sim \mathcal{X}\times \mathcal{Y}$ and $r$ denotes the learned splitting rule consisting of a splitting feature and a splitting value. 

In centralized training, the computation of function $h$ can be completed without communication since both features $x$ and labels $y$ are owned by a single party locally. 
However, in the VFL scenario, features can be distributed among multiple parties, and labels are only kept in one task party. Both features and labels could not be shared directly due to privacy concerns.

To complete the training process in the VFL scenario, the computation of $h$ is transformed into several sub-tasks. These sub-tasks would be assigned to different parties accordingly. The results of these sub-tasks would be aggregated, after being protected to avoid privacy leakage if necessary, to learn the optimal splitting rule at each internal node.
Previous studies on TBMs in the VFL scenario focus on how to define the sub-tasks and how to protect the exchanged information, which can be divided into two categories according to their computation and communication protocols, \ie feature-gathering TBMs and label-scattering TBMs.

\begin{figure*}[t]
\centering  

\subfigure[Characteristics of feature-gathering TBMs.]{ 
\includegraphics[width=0.75\textwidth]{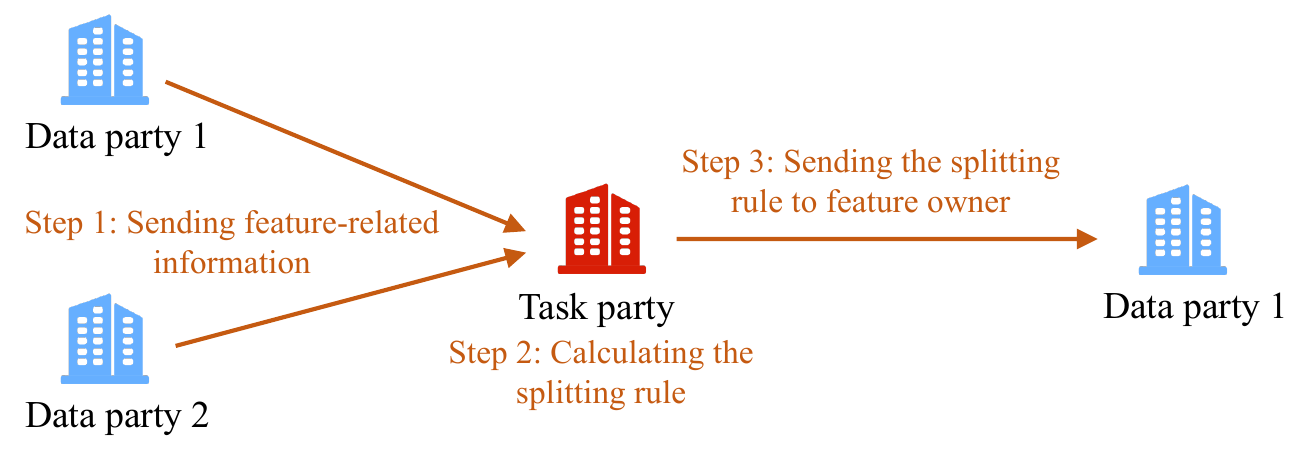}
}

\subfigure[Characteristics of label-scattering TBMs.]{  
\includegraphics[width=0.95\textwidth]{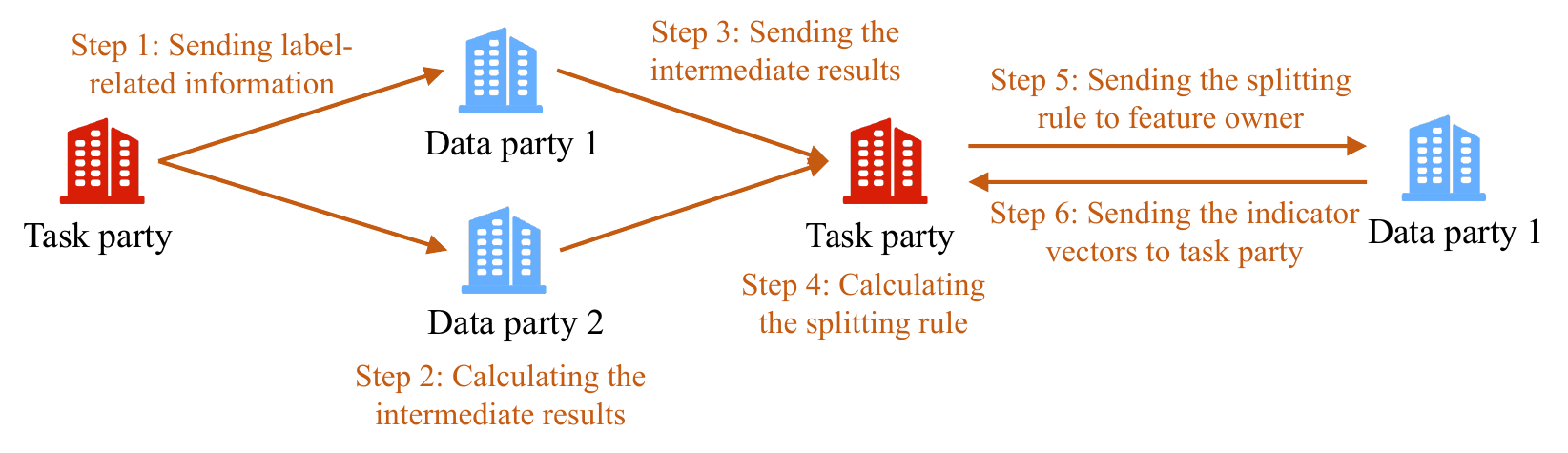}  
}

\caption{Comparisons between feature-gathering and label-scattering TBMs.\label{fig: feature-gathering and label-scattering}}     
\end{figure*}

Figure~\ref{fig: feature-gathering and label-scattering} illustrates the characteristics of feature-gathering and label-scattering TBMs. We will provide a more detailed introduction and comparison below.

\subsection{Feature-gathering TBMs}
\label{subsec:feature-gathering model}
\subsubsection{Splitting rule finding}
The main idea of feature-gathering TBMs is to modify the splitting rule finding function $h$ as:
\begin{equation}
	r=h_\text{feature}\left[g(x^{(1)}), g(x^{(2)}),\ldots,  g(x^{(M)}), y\right], \label{eq:split_funcs_order_based}
\end{equation}
where $x^{(m)} \sim \mathcal{X}_m \ \forall m \in [M]$. In other words, the calculation of splitting rule $r$ contains the following two steps: 
(1) Each data party completes the sub-task functioned as $g$, taking its feature stored locally as input and outputting some (protected) intermediate results $g(x^{(m)})$. Then these intermediate results are sent to the task party; 
(2) After receiving all the intermediate results from the data parties, the task party calculates the splitting rule $r$ based on $g(x^{(m)}) \ \forall m \in [M]$ and the labels $y$.

It is worth pointing out that the splitting rule finding function $h$ defined in Eq.\eqref{eq:split_funcs_order_based} brings some issues for building feature-gathering TBMs. Since the task party only receives the intermediate results rather than the features from data parties, it could not identify the found splitting rules (\ie the splitting features and splitting values) without communicating with the owner of the splitting features. 

To solve this, what the task party actually obtained from $h_\text{feature}$ is a ``pointer'' of the splitting feature and splitting value, and the ``pointer'' should be sent back to the corresponding data party that is able to look up the real splitting feature and splitting value based on the ``pointer''. 
For example, the splitting rule found by the task party can be ``the $5$-th feature of party $m$, the $120$-th-ranked value'', and it can only be recovered by party $m$ to ``age'' (\ie the splitting feature) and ``45'' (\ie the splitting value).
The splitting features and splitting values are stored in data parties, and the task party only knows the ``pointer'' and should query for these splitting results when needed, such as during the inference process (more details can be found in Section~\ref{subsec:inference procedure}).

One of the widely adopted instantiations of the function $g$ is getting the ordinal numbers of samples according to one's features, \aka the data sample indices ranked by the feature values. The data parties sort their data according to each of the features, respectively, and send the resulting ordinal numbers to the task party. 
Therefore, based on these ordinal numbers, the task party can calculate the gains achieved by different splitting rules and find the best one.

\subsubsection{Tree building}
The overall process of building feature-gathering TBMs is demonstrated in Algorithm~\ref{algo: order_based_example}. Specifically, the task party builds $T$ decision trees based on the received intermediate results $g(x^{(m)})\ \forall m \in [M]$ and labels $y$. 

\begin{algorithm}[t]
	\caption{Trees building of feature-gathering model} \label{algo: order_based_example}
	\begin{algorithmic}[1]
		\FOR{each data party $m \in [M]$}
		\STATE Calculate the intermediate results $g(x^{(m)})$;
		\STATE Send $g(x^{(m)})$ to the task party;
		\ENDFOR
  
		\FOR{tree number $t=1, 2, \ldots , T$, the task party}
		\STATE Initialization: $N=\emptyset$, which consists of 2-term tuples. Each tuple contains a node from a tree as its first element, and the corresponding node data as its second element,
  
        \ \ data sample $D_\text{entire}=\{g(x^{(1)}), g(x^{(2)}), \ldots, g(x^{(m)}), y\}$;
		\STATE Build a tree from root: $N \gets N \cup \{ (root\_node, D_\text{entire}) \}$;
		\WHILE{$N \neq \emptyset$}
		\STATE Get $(n, D)$ from $N$ to be traversed;
		\IF{$n$ is a {\it leaf node}}
		\STATE Set the output of $n$;
		\ELSE
		\STATE Find the splitting rule achieving the best gain:
        
        \ \ $r=h_\text{feature}(D)$ ;
		\STATE Send $r$ to the corresponding data party;
		\STATE Split $D$ into $D_\text{left}$ and $D_\text{right}$ according to $r$;
		\STATE Add $n$'s children (denoted as $n_\text{left}$ and $n_\text{right}$) to $N$: 
  
        \ \ $N \gets N \cup \{(n_\text{left}, D_\text{left}), (n_\text{right}, D_\text{right})\}$;
		\ENDIF
		\STATE Remove the traversed node: $N \gets N \backslash \{(n,D)\}$;
		\ENDWHILE
		\ENDFOR
	\end{algorithmic}
\end{algorithm}

For each decision tree, the data party traverses from the root node to the leaf nodes, finding the splitting rule for each internal node that can achieve the best gain on the data samples. 
All the data samples are fed into the root of the tree at the beginning (line 6),  and partitioned into two subsets (or $k$ subsets for a $k$-way tree) according to the splitting rule (line 15). 
These two subsets go to the children of the root, respectively (line 16), and such a partition process is repeated at every traversed node until the leaf nodes are reached. 
For the leaf node, the task party sets the output of the leaf node according to the reached data (line 11), \eg calculating a weight according to Equation~\eqref{eq:xgb_set_weight} in XGBoost or performing major voting in random forest. 
The naive approach for finding the best splitting rule, \ie $h_\text{feature}$  in line 13, can be exhaustive enumeration: For each possible splitting position, the task party calculates the gain (\eg the score defined in Equation~\eqref{eq:xgb_gain} in XGBoost or the Gini coefficient in a random forest) accordingly.
The task party finally chooses the splitting rule that can achieve the best gain at every internal node. 
Several advanced algorithms for balancing the efficiency and effectiveness of finding the optimal splitting rule have been proposed recently~\cite{ke2017lightgbm, dorogush2018catboost}.

In a nutshell, we can conclude the communication and computation protocol of feature-gathering TBMs: (1) Feature-related information is sent from data parties to the task party in a privacy-preserving manner; (2) Most of the computation behaviors happen at the task party, including calculating the gain of different splitting rules to find the solution (\ie line 13 in Algorithm~\ref{algo: order_based_example}), and further partitioning data samples into subsets (\ie line 15 in Algorithm~\ref{algo: order_based_example}). 
As a result, it can be implied that the task party can be the bottleneck of computation and communication in feature-gathering TBMs. The reason is that both feature-related information and labels are pooled at the task party for building decision trees.

\begin{figure}
    \centering
    \includegraphics[width=0.76\textwidth]{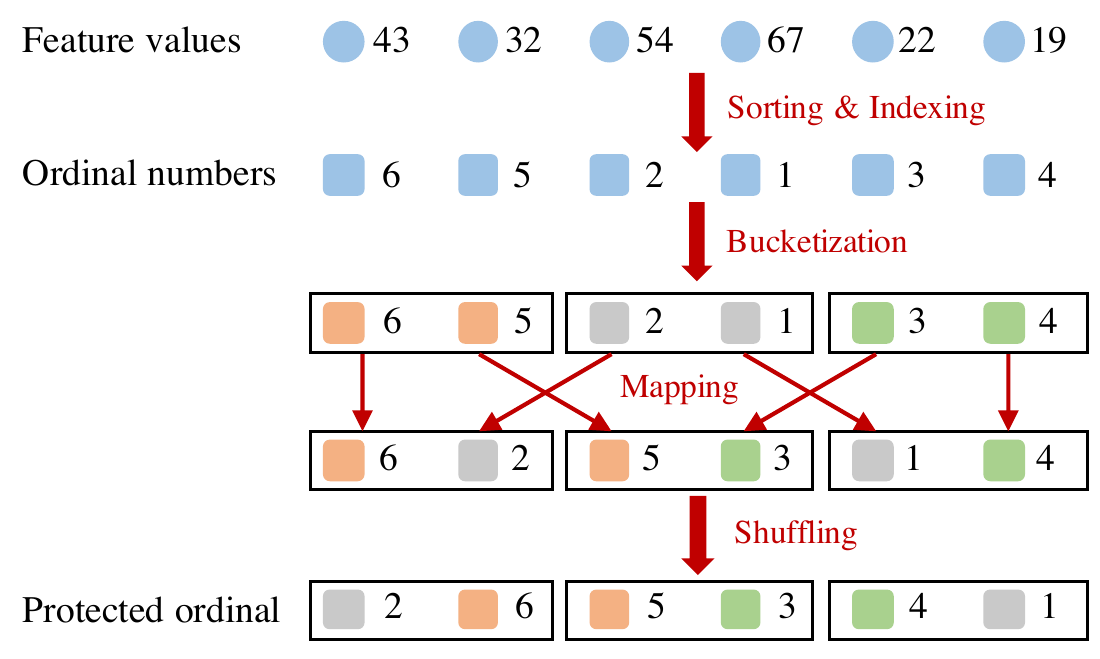}
    \caption{Illustration of the privacy protection algorithms proposed by FederBoost~\cite{tian2020federboost}.\label{fig:dp picture}}
\end{figure}

\subsubsection{Privacy protection mechanisms.}
\label{subset:feature_basd_proetction}
The privacy threats, aimed at the shared feature-related information (\eg the ordinal numbers of data samples) in feature-gathering TBMs, come from the semi-honest task part and insecure communication channels.
In order to avoid leaking private information, it is necessary for the data parties to apply some protection mechanisms to feature-related information before sharing it. 
Such protection mechanisms should be carefully designed to balance the strength of protection and the informativeness of shared information. 
Here we briefly introduce several representative algorithms.

FederBoost~\cite{tian2020federboost} proposes to provide LDP privacy protection for ordinal numbers by adding noises to $g(x^{(m)})$, \ie making improvements at line $2$ in Algorithm \ref{algo: order_based_example}.
To be more specific, as shown in Figure~\ref{fig:dp picture}, each data party first sorts the training samples according to its feature values for obtaining the ordinal numbers, and then partitions the ordinal numbers into several buckets sequentially.
To achieve $\epsilon$-DP, a mapping is applied, making that each ordinal number might move to other buckets with a certain probability (controlled by a hyperparameter to achieve a good utility-privacy trade-off), or just stay at the correct bucket.
The ordinal numbers inside the buckets would be randomly shuffled before sharing, therefore the relative order between buckets can be preserved mostly while the orders within each bucket are protected.
In this way, the data parties generate the protected ordinal numbers that can be sent to the task party for training feature-gathering TBMs.

\begin{figure}
    \centering
    \includegraphics[width=0.76\textwidth]{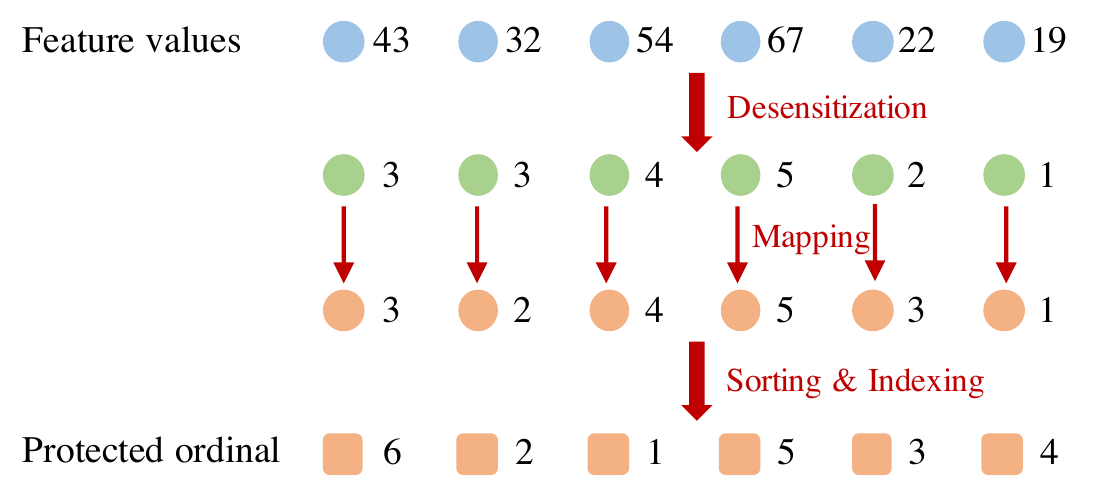}
    \caption{Illustration of the privacy protection algorithms proposed by OpBoost~\cite{li2022opboost}.\label{fig:dLDP picture}}
\end{figure}

OpBoost~\cite{li2022opboost} designs a probabilistic order-preserving desensitization algorithm for privacy-preserving vertical federated tree boosting, which satisfies distance-based LDP.
The main idea of OpBoost is to pay more attention to enhancing the indistinguishability of private feature values from their nearby neighbors.
Indeed, as shown in Figure~\ref{fig:dLDP picture}, the feature values are desensitized into a unified discrete value domain with the predefined lower and upper bound. After that, a mapping satisfying distance-based LDP is used to transform each feature value in the domain based on a distance-based scoring function, \ie values will be mapped to nearby values with high probability.
Finally, the data parties sort the training samples according to these desensitized feature values, generating the protected ordinal numbers.
Such protected ordinal numbers are sent to the task party for training feature-gathering TBMs, which is proven to be a good balance of preventing privacy leakage and preserving useful information.

\subsubsection{Summary.}
In a nutshell, feature-gathering TBMs propose that the data parties calculate ordinal numbers of samples based on their private feature values in a private-preserving manner, and send these desensitized results to the task party for learning optimal splitting rules.
As a result, the task party gathers the feature-related information from all data parties, and assumes the responsibility for the computation of finding the optimal splitting rules at every node of the decision trees, while the data parties only need to sort the samples and disrupt the generated ordinal numbers.
The major communication overhead of feature-gathering TBMs is the ordinal numbers of samples, which would be sent from the data parties to the task party. The characteristics of feature-gathering TBMs are summarized in Table~\ref{table: characteristics and comparisons}.

Note that there is a trade-off between model utility and privacy protection strength when training feature-gathering TBMs, as pointed out by previous studies~\cite{tian2020federboost, li2022opboost}.
Though differential privacy algorithms can enhance privacy protection strength, they might also cause a slight performance drop in the learned model.
Such a trade-off implies that the protection strength should be carefully determined in real-world applications, and also inspires the research community to design advanced algorithms to achieve better model utility with a certain allocated privacy budget.

\begin{table*}[t]
\caption{The characteristics of feature-gathering TBMs and label-scattering TBMs. \label{table: characteristics and comparisons}}
\centering
\resizebox{0.95\columnwidth}{!}{
\begin{tabular}{lcc}
   \toprule
    & {\bf Feature-gathering TBMs} & {\bf Label-scattering TBMs} \\ 
   \midrule
   \multirow{2}{*}{Shared information} & Feature-related information & Label-related information \\
    & (from data parties to task party) & (from task party to data parties) \\
   Computation contributors & Task party & Data parties and task party \\
   Privacy protection & Differential privacy &  Homomorphic encryption and secret sharing \\
    Related studies &\cite{tian2020federboost, li2022opboost}  & \cite{cheng2021secureboost, wu2020privacy, chen2021fed, han2022fedgbf, fang2021large, xie2022efficient, liu2020federated, song2021federated}  \\
  \bottomrule
\end{tabular}
}
\end{table*}

\subsection{Label-scattering TBMs}
\subsubsection{Splitting rule finding.} 
Different from feature-gathering TBMs introduced above, label-scattering TBMs propose another protocol for building decision trees in the VFL scenario, \ie the task party broadcasts the label-related information to data parties.

Formally, the main idea of label-scattering TBMs is to modify the splitting rule finding function $h$ as:
\begin{equation}
	r=h_\text{label}\left[g(x^{(1)}, e(y)), g(x^{(2)}, e(y)),\ldots,  g(x^{(M)}, e(y))\right], \label{eq:split_funcs_label_based}
\end{equation}
where $x^{(m)} \sim \mathcal{X}_m \ \forall m \in [M]$, and $e(y)$ denotes the label-related information calculated by the task party via the function $e$. Accordingly, the calculation of the splitting rule consists of the following three steps: 
(1) The task party applies the function $e$ on label $y$, and broadcasts the produced label-related information $e(y)$ to all data parties; 
(2) After receiving $e(y)$, each data party completes the sub-task functioned as $g$ based on $e(y)$ and its private feature values, and sends the intermediate results $g(x^{(m)}, e(y))$ back to the task party; 
(3) The task party calculates the splitting rule $r$ based on the intermediate results received from data parties.

Similar to that in feature-gathering TBMs, the obtained $h_\text{label}$ is just a ``pointer'' of the splitting feature and splitting value, which should be sent to the corresponding data party for querying. 
One instantiation of $e(y)$ can be the first-order and second-order gradients, and $g$ can be sorting the received $e(y)$ based on the ordinal numbers of samples ranked by data parties' feature values.

\begin{algorithm}[t]
	\caption{Trees building of label-scattering model} \label{algo: label_sacttering_example}
	\begin{algorithmic}[1]
		\FOR{tree number $t=1, 2, \ldots , T$}
            \STATE  The task party calculates $e(y)$;
            \STATE  The task party broadcasts $e(y)$ to all the data parties;
            \FOR{each data party $m \in [M]$}
		\STATE Calculate the intermediate results $g(x^{(m)}, e(y))$;
		\STATE Send $g(x^{(m)}, e(y))$ to the task party;
		\ENDFOR
		\STATE After receiving intermediate results, the task party performs initialization: $N=\emptyset$, which consists of 2-term tuples. Each tuple contains a node from a tree as its first element, and the corresponding node data as its second element,
  
        \ \ data sample $D_\text{entire}=\{g(x^{(1)}, e(y)), \ldots, g(x^{(m)}, e(y))\}$;
		\STATE The task party builds a tree from the root: 
  
        \ \ $N \gets N \cup \{ (root\_node, D_\text{entire}) \}$;
		\WHILE{$N \neq \emptyset$, the task party}
		\STATE Get $(n, D)$ from $N$ to be traversed;
		\IF{$n$ is a {\it leaf node}}
		\STATE Set the output of $n$;
		\ELSE
		\STATE Find the splitting rule achieving the best gain:
  
        \ \ $r=h_\text{label}(D)$;
		\STATE Send $r$ to the corresponding data party;
        \STATE After receiving $r$, the data party splits $D$ into $D_\text{left}$ and $D_\text{right}$ and sends indicator vectors back;
        \STATE Add $n$'s children (denoted as $n_\text{left}$ and $n_\text{right}$) to $N$: 
        
        \ \ $N \gets N \cup \{(n_\text{left}, D_\text{left}), (n_\text{right}, D_\text{right})\}$;
		\ENDIF
		\STATE Remove the traversed node: $N \gets N \backslash \{(n,D)\}$;
		\ENDWHILE
		\ENDFOR
	\end{algorithmic}
\end{algorithm}

\subsubsection{Tree building.}
We describe the overall process of training label-scattering TBMs in Algorithm~\ref{algo: label_sacttering_example}. 
The goal is to build $T$ decision trees collaboratively. 

For each decision tree, the data party needs to generate the label-related information $e(y)$ and then broadcast $e(y)$ to all data parties (lines 2-3).
In most cases, $e(y)$ can be different when building different decision trees, \eg when $e(y)$ is the set of first-order and second-order gradients, it relies on both labels $y$ and decision trees built previously.
For each data party $m\in [M]$ who has received $e(y)$, it calculates the intermediate results and sends the results back to the task party (lines 5-6).
After receiving the intermediate results from all data parties, the task party is able to traverse from the root node to the leaf nodes for finding the splitting rules that achieve the best gain on the reached data samples (lines 9-21). Such a traversing process is almost the same as the process in feature-gathering TBMs, except for how to partition the data samples into several subsets (line 18). In label-scattering TBMs, the task party could not partition the data samples into subsets based on the found splitting rule $r$, since the task party does not have the feature-related information. Therefore, the task party has to send $r$ to the corresponding data party and wait for the feedback, including some (protected) indicator vectors to denote the data subsets attached to the children of the traversed node.

From the algorithm, we can conclude that in label-scattering TBMs, (1) The task party broadcasts the label-related information to data parties, which might happen before building every decision tree if such information would update; (2) Both the task party and data parties involve in the process of tree building, \ie the traversing process from roots to leaf nodes. Specifically, keeping communication with each other, the task party finds the splitting rules, and the data parties partition the data samples into subsets accordingly.
Compared to feature-gathering TBMs, the communication between the task party and the data parties in label-scattering TBMs is more frequent.

\subsubsection{Privacy protection mechanisms.} 
In label-scattering TBMs, some information is vulnerable and required for further protection, including the label-related information $e(y)$, the intermediate results $g(x^{(m)}, e(y))$, and the indicator vectors. In order to provide protection, researchers propose several advanced privacy protection mechanisms based on homomorphic encryption and secure multi-party computation.

\begin{figure}[t]
    \centering
    \includegraphics[width=0.95\textwidth]{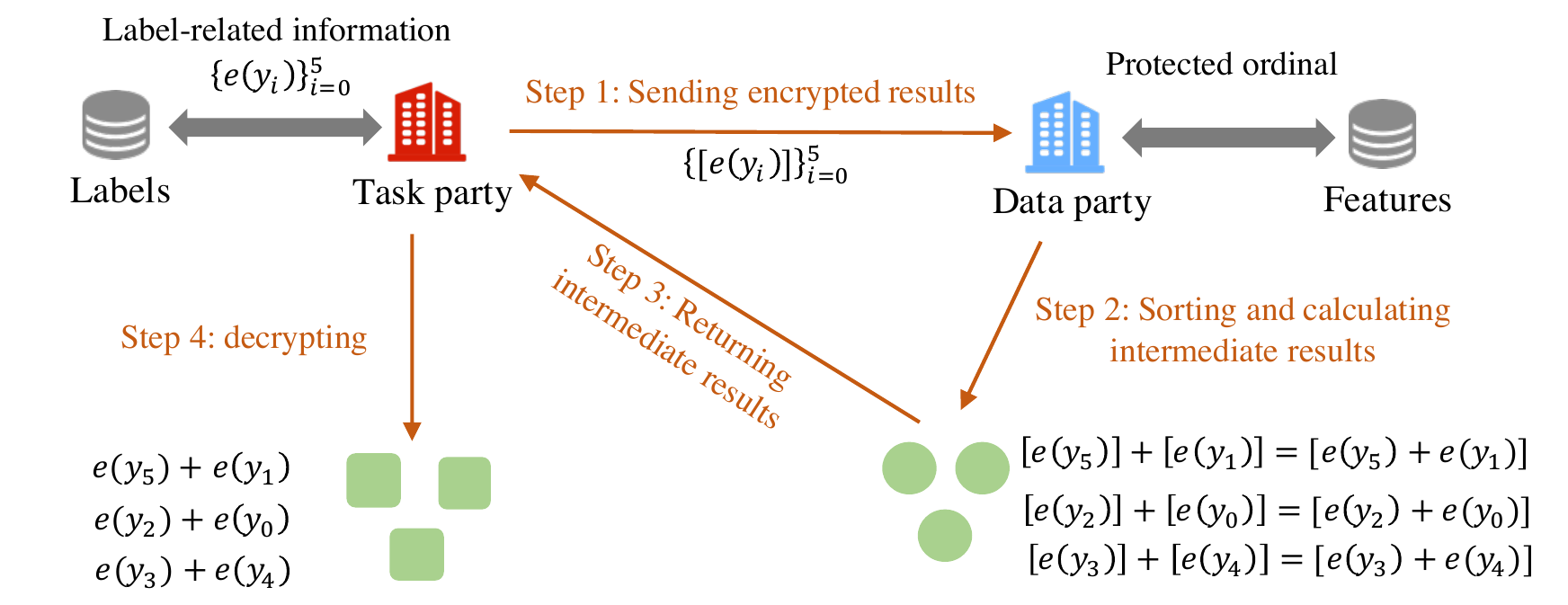}
    \caption{Illustration of SecureBoost~\cite{cheng2021secureboost}.}
    \label{fig: PHE partial sum picture}
\end{figure}

SecureBoost~\cite{cheng2021secureboost} is focused on XGBoost and applies homomorphic encryption algorithms on $e(y)$ for enhancing privacy protection. 
An illustration of SecureBoost is shown in Figure~\ref{fig: PHE partial sum picture}. Specifically, the task party broadcasts the encrypted $e(y)$ to all data parties. Each data party calculates $g(x^{(m)}, e(y))$ independently, following the process of sorting $e(y)$ based on the ordinal numbers ranked by its feature values and partitioning the sorted $e(y)$ into several buckets.
These values inside each bucket are summed up and finally sent back to the task party. Therefore, the task party can perform decryption to recover the sum of values in each bucket (due to the characteristic of homomorphic encryption) but cannot get the individual values of $e(y)$. 

Further, with the aim of protecting both the label-related information, the intermediate results, and the indicator vectors, recent studies~\cite{wu2020privacy, fang2021large} propose to combine partially homomorphic encryption and additive secret sharing techniques.
To be more specific, for the task party, each node is attached to an indicator vector for training samples, where each element is a binary value to represent whether a sample reaches the node (value $1$) or not (value $0$). The indicator vector at the root node is initialized as $1$s.

\begin{figure}
    \centering
    \includegraphics[width=0.95\textwidth]{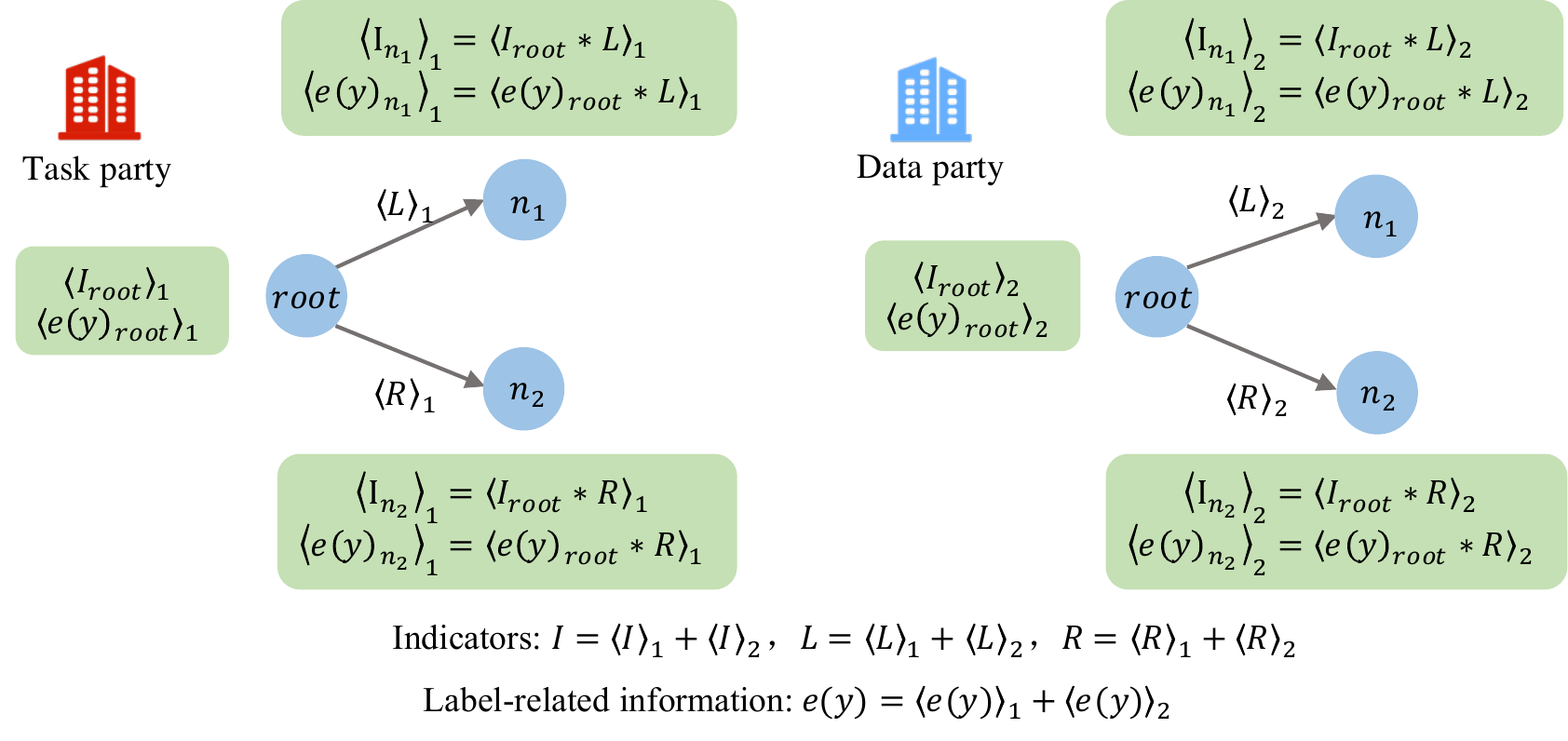}
    \caption{An example of applying secret sharing in label-scattering model.\label{fig: ss picture}}
\end{figure}

Then the task party splits each indicator vector into several frames, and sends one of the frames to a data party. 
And the task party also sends the encrypted label-related information $e(y)$.
After receiving these results, each data party sorts $e(y)$ according to the ordinal numbers ranked by its feature values, splits the sorted $e(y)$ into several frames, and sends one of the frames to a data/task party.
Thus each party owns one frame of the encrypted $e(y)$ with different orders, and they could jointly perform the computation to get the maximal splitting gain for learning the optimal splitting rules via the secret sharing division technique.
The data party that holds the splitting feature can update the indicator to represent which subtree would be traversed according to learned splitting rules for each training sample.
The updated indicators are also split into several frames and sent to different parties, which can be used for further computation via secret sharing multiplication. In Figure~\ref{fig: ss picture} we show the splitting process among two parties, where each party only holds one frame of indicators and label-related information. 
Here, $L$ and $R$ mean the indicator vector of the incidences of samples belonging to the left and right subtree, respectively. By secret sharing multiplication, the children's nodes also hold the frames of the protected information.

Although applying partially homomorphic encryption and additive secret sharing techniques can enhance the privacy protection strength for label-scattering TBMs, the additional computation and computation cost they might bring is non-negligible in real-world VFL applications.

\subsubsection{Summary.}
Generally speaking, label-scattering TBMs suggest that the task party broadcasts the label-related information to all data parties without raising privacy issues, and data parties calculate and then send the intermediate results back to the task party for learning splitting rules.
Finally, the splitting rules would be sent to the data party that owns the corresponding splitting feature for generating the partition of data samples.
As a result, the task party scatters the label-related information to data parties, and both the data parties and the task party make their computation for building label-scattering TBMs.

The major communication overhead of label-scattering TBMs is the label-related information, which might be broadcast by the task party before building every decision tree if such information would be updated after finishing building a tree.
The characteristics of label-scattering TBMs are summarized in Table~\ref{table: characteristics and comparisons}, which also shows the comparison between feature-gathering TBMs and label-scattering TBMs to highlight their differences.

\subsection{Inference Procedure}
\label{subsec:inference procedure}
In the previous sections, we introduce the training process of two types of TBMs in the VFL scenario. In this section, we describe their inference procedure, which also needs collaboration among multiple parties since the learned models are decentralized.

There exist two inference frameworks for TBMs in the VFL scenario, distinguished by whether the inference procedure is led by the task party or accomplished through multiple parties. Specifically, to produce a predicted result, a test sample traverses from the root node to one leaf node for each built decision tree, following the path generated according to the learned splitting rules at each internal node. 
One inference framework~\cite{cheng2021secureboost, tian2020federboost} proposes the inference procedure can be led by the task party, since it holds the learned splitting rules (\ie the ``pointers'') at every internal node. When a test sample reaches an internal node, the task party sends a request to the corresponding data party (which is the owner of the splitting feature) to query which subtree the sample should choose to go.
It can be implied that the communication between the task party and the data party aforementioned might expose the splitting features and values during such querying operations, which motivates researchers~\cite{fang2021large} to provide protection via secret sharing techniques.

\begin{figure}[t]
    \centering
    \includegraphics[width=0.65\textwidth]{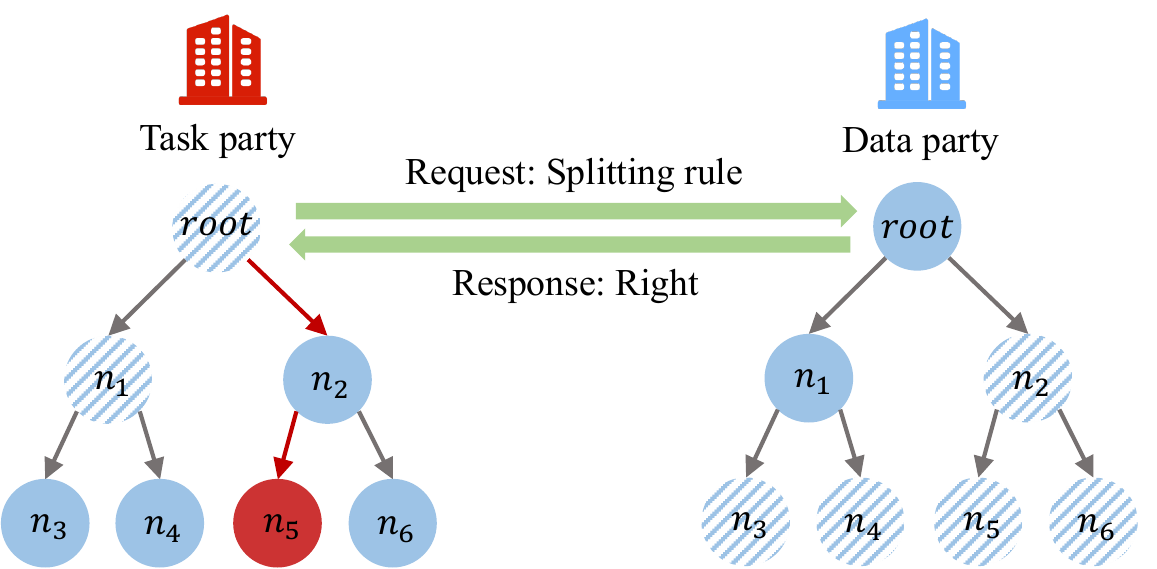}
    \caption{Illustration of the inference procedure led by the task party.}
    \label{fig: task party inference}
\end{figure}

Figure~\ref{fig: task party inference} illustrates an example of the inference procedure led by the task party, where solid circles represent nodes that hold the splitting rules or values, while dashed circles represent nodes that do not.
We assume the test case reaches the leaf node $n_5$, and the inference procedure can be summarized as follows. The inference begins at the root node. Since the task party does not hold the splitting rule at the root node, it needs to request the splitting rule associated with the root node from the data party. When receiving the request, the data party responds with an indicator ``right'' (which may be encrypted) to the task party. The inference procedure then moves on to node $n_5$. Based on the splitting rule held by the task party, the test case finally arrives at the corresponding leaf node.

\begin{figure}[t]
    \centering
    \includegraphics[width=0.85\textwidth]{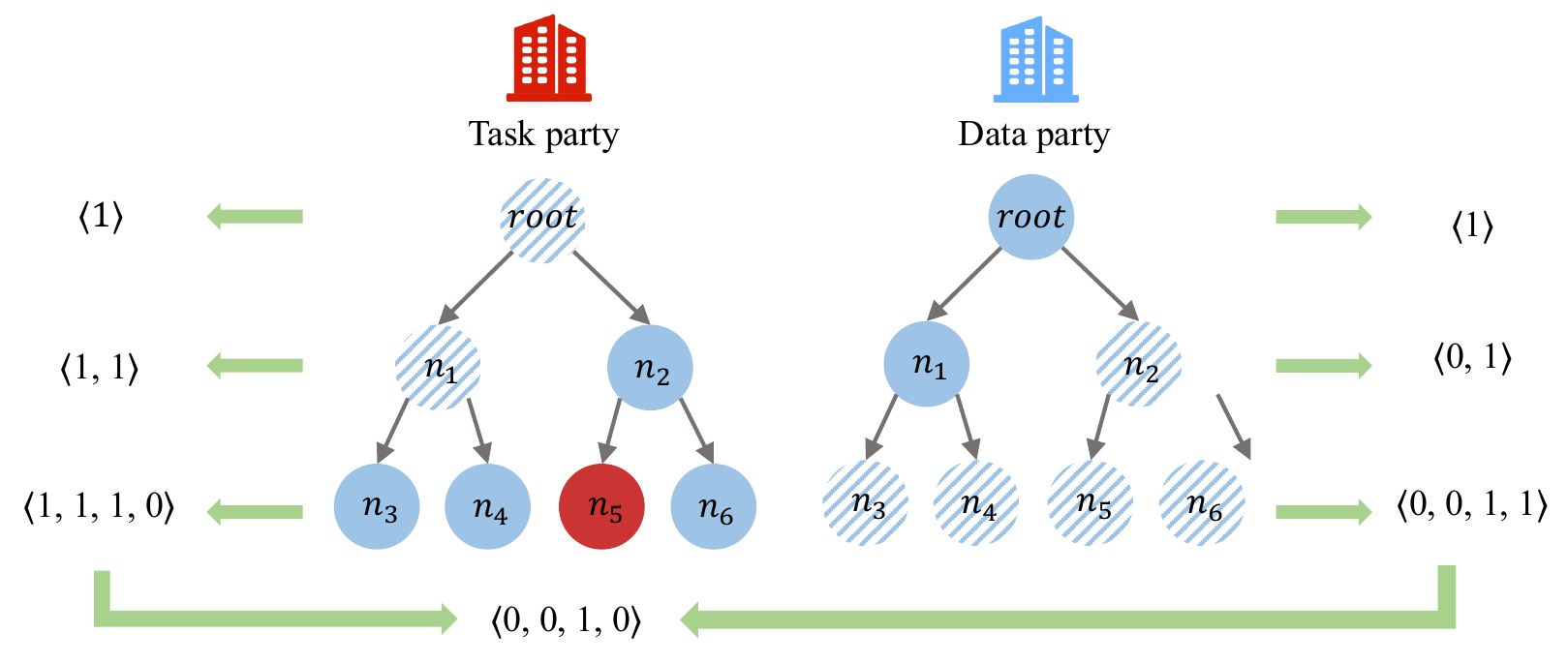}
    \caption{Illustration of the inference procedure accomplished through multiple parties.}
    \label{fig: incidence vector inference}
\end{figure}

Another inference framework~\cite{wu2020privacy, chen2021fed, yao2022efficient} needs to be accomplished by multiple parties. 
Taking the case of binary trees as an example, each party starts at the root node and holds a binary indicator valued as $1$ to denote that the test sample reaches the node. 
Then, if the party happens to hold the splitting feature and splitting value at the node, it knows which subtree the test sample should choose to go, producing $\langle1,0\rangle$ (for going to the left subtree) or $\langle0,1\rangle$ (for going to the right subtree) accordingly. 
If the party has no information about the splitting rules, it produces $\langle1,1\rangle$ (the test sample might go to any one of the subtrees) or $\langle0,0\rangle$ (the test sample has not chosen this path) following the indicator value at the current node (\ie their parent node).
Thus, each party can produce an indicator vector whose length is the same as the number of leaf nodes, which can be gathered by performing element-wise {\sc AND} operation to know which leaf node the test sample finally reached.
To further avoid privacy leakage brought by sharing the indicator vectors, researchers~\cite{wu2020privacy, chen2021fed} propose to apply homomorphic encryption algorithms to mask the indicator vectors.

Figure~\ref{fig: incidence vector inference} illustrates an example of the inference procedure accomplished through multiple parties, where solid circles represent nodes that hold the splitting rules or values, while dashed circles indicate nodes that do not.
We assume the test case reaches the leaf node $n_5$, and the inference procedure can be summarized as follows. As the initialization, both parties start with indicators valued as $\langle 1\rangle$. The data party then updates the indicator to $\langle0, 1\rangle$ according to the splitting rules, indicating that the test case would take the right branch. Meanwhile, the task party updates its indicator to $\langle1, 1\rangle$ since it lacks knowledge of the splitting rule at this point. As the inference moves to the next depth level, the data party updates the indicator from 0 to $\langle0, 0\rangle$ for $n_1$ (although the data party holds the splitting rule of $n_1$, the test case does not choose this branch), and updates the indicator from 1 to $\langle1, 1\rangle$ for $n_2$ (as it does not hold the splitting rule). This results in concatenating the indicators to $\langle0, 0, 1, 1\rangle$. On the other hand, the task party updates the indicator from 1 to $\langle1, 1\rangle$ for $n_1$ (as it does not hold the splitting rule), and updates the indicator from 1 to $\langle1, 0\rangle$ for $n_2$ based on the calculation from the splitting rule. This results in concatenating the indicators to $\langle1, 1, 1, 0\rangle$. Finally, all indicators are aggregated using an element-wise logical {\sc AND} operation, which produces $\langle0, 0, 1, 0\rangle$ and indicates that the test case finally arrives at leaf node $n_5$.

\section{Infrastructure}
\label{sec:infrastructure}
In this section, we review open-source FL platforms for supporting tree-based models (TBMs) in the vertical federated learning (VFL) scenario, and further summarize several principles to promote the design of infrastructure.

\subsection{FL Platforms}
Remarkable progress has been made by open-source FL platforms~\cite{liu2021fate, fedtree, xie2022federatedscope, fedml, fedscale, flower, wang2022federatedscope, kuang2024fsllm} in supporting users to conveniently apply FL in real-world applications and developing new FL algorithms, which cover various FL scenarios.
In order to support TBMs in the VFL scenario, FL platforms are expected to allow multiple parties to execute various kinds of subroutines for receiving, handling, and sending different types of information and completing different computation tasks.

Inspired by previous studies~\cite{liu2022unifed, yang2019federated, aledhari2020federated, li2021survey, lim2020federated}, we briefly summarize the open-source FL platforms that can satisfy the requirements of applying TBMs in the VFL scenario, including:
\begin{itemize}
    \item \textbf{FATE}\footnote{https://github.com/FederatedAI/FATE} is an industrial-grade FL platform that supports the secure computation of different kinds of machine learning algorithms, including several tree-based models such as GBDT.
    \item \textbf{Fedlearner}\footnote{https://github.com/bytedance/fedlearner} is focused on multi-party collaborative tasks, and provides the implementation of SecureBoost for the VFL scenario.
    \item \textbf{FedTree}\footnote{https://github.com/Xtra-Computing/FedTree} is a specially designed FL platform targeting tree-based models, which provides different types of tree-based models such as GBDT and random forests, equipped with various privacy protection mechanisms.
    \item \textbf{SecretFlow}\footnote{https://github.com/secretflow/secretflow} provides device abstraction for conveniently applying SMPC and releases the implementation of label-scattering TBMs, \eg HEP-XGB and CRP-XGB.
    \item \textbf{FS-Tree}\footnote{https://github.com/alibaba/FederatedScope/tree/master/federatedscope/vertical\_fl} is a module in FederatedScope~\cite{xie2022federatedscope} designed for TBMs, which is a flexible and easy-to-use FL platform based on an event-driven architecture. It includes both feature-gathering TBMs and label-scattering TBMs, and provides rich types of privacy protection mechanisms as plugins.
\end{itemize}

\subsection{Design Principles}
Considering the diversity of VFL applications, we summarize the design principles of FL platforms to better satisfy the requirements of applying TBMs in the VFL scenario:
\begin{itemize}
    \item {\it Flexibility}. Flexibility refers to the platform's ability to support various types of communication and computation protocols when applying TBMs in the VFL scenario. Specifically, a platform exhibiting strong flexibility should effectively handle the actions associated with existing types of communication and computation protocols, including sending, receiving, and processing behaviors during both training and inference procedures. In contrast, platforms that only support a specific protocol or lack capabilities for diverse actions are considered less flexible.
    \item {\it Extensibility}. One of the main targets of constructing FL platforms is to save the effort of developers in implementing new algorithms, which motivates FL platforms to be extendable. Extensibility denotes the extent to which a platform can conveniently support the addition of new modules, which is particularly crucial for the ongoing development of TBMs. When the modules provided by the platform are tightly coupled, it increases the effort required for developers to introduce new modules or enhance existing ones, thereby diminishing the extensibility of platforms. A design featuring pluggable modules can significantly enhance extensibility. The level of extensibility of a platform can be assessed by measuring the effort required for developers to implement a reasonable extension, such as the number of lines of code added or the number of modified files.
    \item {\it Scalability}. Scalability has two dimensions within the context of TBMs in VFL: the platform's ability to handle increasing data volumes (e.g., additional features and samples) and its capacity to support an increasing number of participants. A platform with strong scalability should ensure that, as data volumes or participants increase, its efficiency does not exceed linear growth under specific computational resource conditions. 
    \item {\it Security}. Privacy protection is particularly important for applying TBMs in the VFL scenario, since feature-related and label-related information is required to be exchanged, which are more vulnerable than those in the HFL scenario (\eg the model parameters). Security of the platform primarily pertains to the availability of advanced privacy protection algorithms that can be effectively implemented to address the varied requirements of different real-world applications. For TBMs in VFL, different protocols might necessitate different privacy protection algorithms. Consequently, platforms that facilitate user-friendly implementation and allow for parameter adjustments tailored to specific protection needs demonstrate superior security.
\end{itemize}

Overall, the aforementioned design principles motivate us to describe different subroutines as separate and pluggable behaviors, and to minimize the dependence between different parties and subroutines.
We hope that these principles can further inspire the community to make improvements on FL platforms in supporting tree-based models.

\section{Experiments}
\label{sec:exp}

In this section, we conduct a series of experiments on several widely-used datasets. We aim to provide an empirical understanding of the characteristics of different types of TBMs, \ie feature-gathering TBMs and label-scattering TBMs. 
Meanwhile, we demonstrate the trade-off among model utility, protection strength, and computation/communication cost, which should be carefully balanced in real-world applications.

\subsection{Datasets and Metrics}

We adopt the following datasets in the experiments:
\begin{itemize}
    \item \textbf{Abalone\footnote{https://archive.ics.uci.edu/ml/datasets/abalone}:} This dataset is prepared for predicting the age of abalone from physical measurements, which contains 4,177 instances and 8 features.
    \item \textbf{Blog\footnote{http://archive.ics.uci.edu/ml/datasets/BlogFeedback}:} The task associated with this dataset is to predict how many comments the post will receive. It contains 52,397 and 7,624 instances for training and testing, respectively, with 280 features in each instance.
    \item \textbf{Adult\footnote{http://archive.ics.uci.edu/ml/datasets/Adult}:} This dataset is prepared for the prediction task to determine whether a person earns over 50K a year or not. It contains 32,561 instances for training, and 16,281 instances for testing, with 14 features. We delete the samples with unknown values, which leads to 30,162 and 15,060 instances for training and testing, respectively.
    \item \textbf{Credit\footnote{https://www.kaggle.com/c/GiveMeSomeCredit/overview}:} It is a credit score dataset that classifies whether a user would suffer from serious financial problems, helping banks determine whether or not a loan should be granted. It contains a total of 150,000 instances and 10 features. 
\end{itemize}

Note that Abalone and Credit datasets do not provide the partition for evaluation, thus for them, we split $80\%$ and $20\%$ of samples for training and evaluation, respectively.
For the evaluation metrics, we adopt Mean Squared Error (MSE) for regression tasks (on Abalone and Blog datasets), and accuracy and Area Under the ROC Curve (AUC) for classification tasks (on Adult and Credit datasets).

\begin{table*}[t]
\caption{The adopted hyperparameters for TBMs.\label{table:hyperparameters}}
\resizebox{0.95\columnwidth}{!}{
\begin{tabular}{llcccccc}
\toprule
Models & Datasets & Number of trees & Depth & Learning rate & $\lambda$ & $\gamma$ & Feature subsample ratio \\ 
\midrule
\multirow{4}{*}{XGBoost}
 & Abalone & 14 & 3 & 0.19  & 0.1 & 0 & 1 \\ 
 & Blog & 15 & 4 & 0.17  & 0.1 & 0 & 1 \\ 
 & Adult & 10 & 3 & 0.56  & 0.1 & 0 & 1 \\ 
 & Credit & 6 & 4 & 0.35  & 0.1 & 0 & 1 \\ 
\midrule
\multirow{4}{*}{GBDT} 
 & Abalone & 15 & 3 & 0.19  & 0.1 & - & 1 \\ 
 & Blog & 15 & 4 & 0.12  & 0.1 & - & 1 \\ 
 & adult & 15 & 4 & 0.49  & 0.1 & - & 1 \\ 
 & Credit & 10 & 4 & 0.1  & 0.1 & - & 1 \\ 
\midrule
\multirow{4}{*}{Random Forest} 
 & Abalone & 10 & 6 & -  & - & - & 1 \\ 
 & Blog & 13 & 6 & -  & - & - & 0.55 \\ 
 & Adult & 10 & 5 & -  & - & - & 0.4 \\ 
 & Credit & 10 & 3 & -  & - & - & 0.2 \\ 
\bottomrule
\end{tabular}}
\end{table*}

\subsection{Implementation Details}

We implement various feature-gathering TBMs and label-scattering TBMs, including RF, GBDT, and XGBoost, based on \ours, a module in FederatedScope~\cite{xie2022federatedscope} designed for TBMs.
The main reason for our choice is that \ours can support flexible information exchange and handling processes with the help of an event-driven architecture. Meanwhile, it allows rich types of privacy protection mechanisms (differential privacy, homomorphic encryption, and secret sharing) as plugins for convenient usage, and has comprehensive benchmarking ability.

In the experiments, we set the number of parties to be $2$, i.e., one task party and one data party. Each party holds part of the features (non-overlap) of the datasets. The categorical features in the datasets have been transformed into numerical types via one-hot encoding, following the settings in previous studies~\cite{li2022opboost}. We set the number of buckets to $50$ for accelerating the process of finding splitting rules. 
We utilize the hyperparameter optimization (HPO) tools provided in FederatedScope for searching the optimal hyperparameters used in various TBMs, and listed the adopted hyperparameters in Table \ref{table:hyperparameters}.

\begin{table*}[t]
\caption{Model performance on four widely-used datasets.\label{table:model_performance}}
\centering
\resizebox{0.95\columnwidth}{!}{
\begin{tabular}{llcccccc}
\toprule
 & \multirow{2}{*}{Models} & Abalone & Blog & \multicolumn{2}{c}{Adult} & \multicolumn{2}{c}{Credit} \\
\cmidrule(lr){5-6} \cmidrule(lr){7-8}
 & & MSE & MSE & Accuracy & AUC & Accuracy & AUC \\ 
\midrule
\multicolumn{1}{c}{\multirow{3}{*}{Feature-gathering}} & Random Forest & 3.866$\pm$0.105 & 576.599$\pm$40.508 & 0.790$\pm$0.064 & 0.836$\pm$0.039 & 0.932$\pm$0.000 & 0.583$\pm$0.039 \\ 
 & GBDT & 4.374$\pm$0.186 & 552.942$\pm$1.637 & 0.840$\pm$0.001 & 0.895$\pm$0.000 & 0.935$\pm$0.000 & 0.812$\pm$0.000 \\ 
 & XGBoost & 4.177$\pm$0.169 & 544.368$\pm$4.423 & 0.844$\pm$0.001 & 0.895$\pm$0.001 & 0.935$\pm$0.002 & 0.823$\pm$0.013 \\ 
\midrule
\multicolumn{1}{c}{\multirow{4}{*}{Label-scattering}} & Random Forest & 3.891$\pm$0.095 & 574.698$\pm$32.156 & 0.823$\pm$0.020  & 0.838$\pm$0.027 & 0.932$\pm$0.000 & 0.566$\pm$ 0.033
\\ 

& GBDT & 4.342$\pm$0.122 & 552.518$\pm$1.959 & 0.840$\pm$0.001 & 0.895$\pm$0.000 & 0.935$\pm$0.000 & 0.812$\pm$0.000 \\ 
& XGBoost & 4.214$\pm$0.100 & 544.429$\pm$4.459 &0.844$\pm$0.001 & 0.896$\pm$0.000 & 0.936$\pm$0.002 & 0.823$\pm$0.011 \\ 
& XGBoost (SS) & 3.588$\pm$0.000 & 546.289$\pm$2.733 & 0.801$\pm$0.003 & 0.820$\pm$0.000  & 0.935$\pm$0.000 & 0.854$\pm$0.001 \\ 
\bottomrule
\end{tabular}}
\end{table*}

\begin{table*}[t]
\caption{Communication frequency on four widely-used datasets\label{table:comm_frequency}}
\vspace{-0.1in}
\centering
\begin{tabular}{llccccc}
\toprule
 & Models & Abalone & Blog & Adult & Credit \\
\midrule
\multicolumn{1}{c}{\multirow{3}{*}{Feature-gathering}} & Random Forest & 133.2$\pm$5.8 & 370.7$\pm$4.7 & 117.9$\pm$7.5 & 37.8$\pm$2.2 \\ 
 & GBDT & 47.1$\pm$1.2 & 127.1$\pm$0.3 & 95.0$\pm$2.2 & 58.0$\pm$0.0 \\ 
 & XGBoost & 43.4$\pm$1.9 & 125.4$\pm$0.7 & 40$\pm$0.0 & 36.1$\pm$0.3 \\ 
\midrule
\multicolumn{1}{c}{\multirow{3}{*}{Label-scattering}} & Random Forest & 743.6$\pm$7.3 & 1,153.6$\pm$10.9 & 410.1$\pm$9.8 & 96.7$\pm$2.8 \\ 
& GBDT & 137.2$\pm$1.5 & 337.1$\pm$0.3 & 303.9$\pm$2.1 & 198.0$\pm$0.0\\
& XGBoost & 127.8$\pm$1.7 & 335.6$\pm$0.8 & 100.0$\pm$0.0 & 119.9$\pm$0.3\\ 
\bottomrule
\end{tabular}
\end{table*}

\subsection{Results and Analysis}
\subsubsection{Comparisons between feature-gathering and label-scattering TBMs}
We conduct a series of experiments to show the differences between feature-gathering TBMs and label-scattering TBMs in terms of model performance and communication frequency, where the communication frequency represents the number of information exchanges between the task party and the data party. 

The experimental results of model performance comparisons are shown in Table~\ref{table:model_performance}, from which we can see that different feature-gathering TBMs and label-scattering TBMs, including random forest, GBDT, and XGBoost, achieve similar model performance in all adopted datasets.
XGBoost (SS) denotes the method proposed in \cite{fang2021large}, combining XGBoost and secret sharing, which is implemented using SecretFlow\footnote{https://github.com/secretflow/secretflow}.
These results are not surprising, since the feature-gathering TBMs and label-scattering TBMs are different in the communication and computation protocols while both of them are reliable to train decision trees well. 
However, these two kinds of TBMs need different communication frequencies to complete the tree-building process, as illustrated in Table~\ref{table:comm_frequency}.
The communication frequency in XGBoost (ss) is meanless, since the method contains a large number of secret sharing computations, including secret sharing addition, division, and comparison, which makes the communication frequency a huge number.
We can observe from the table that label-scattering TBMs need several times of communication frequencies compared to feature-gathering TBMs.
The reason is that, in label-scattering TBMs, the task party needs to broadcast the label-related information to the data parties before building every tree. After finding the splitting rules at each node, the task party and the data parties need more communication for data sample partition compared to that in feature-gathering TBMs.
These experimental results are consistent with the analysis in Section~\ref{sec:method}.

\subsubsection{Model utility versus privacy protection strength}
\begin{figure*}[t]
\centering  
\subfigure[Abalone]{   
\includegraphics[width=0.45\textwidth]{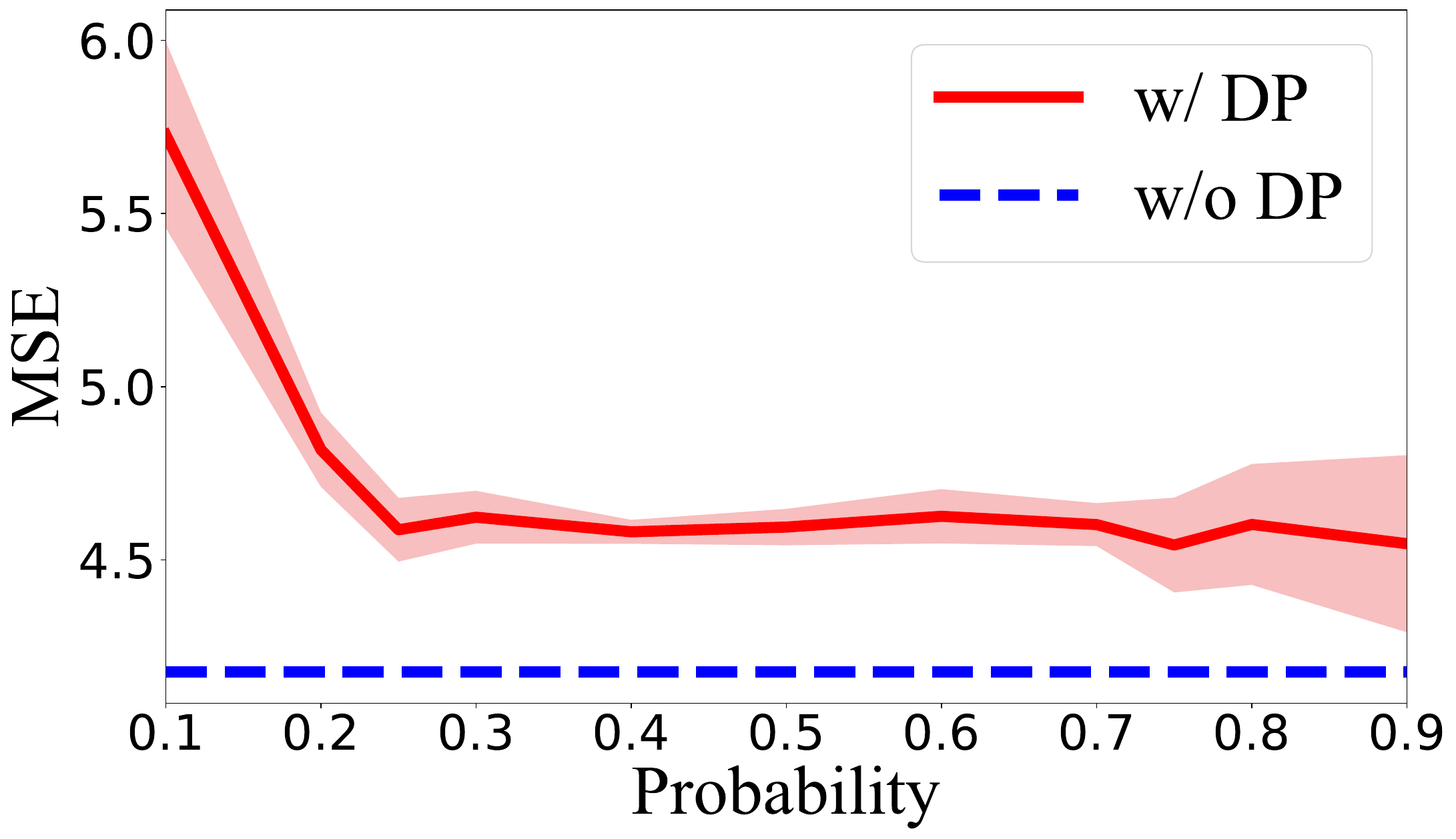} 
}
\subfigure[Blog]{ 
\includegraphics[width=0.45\textwidth]{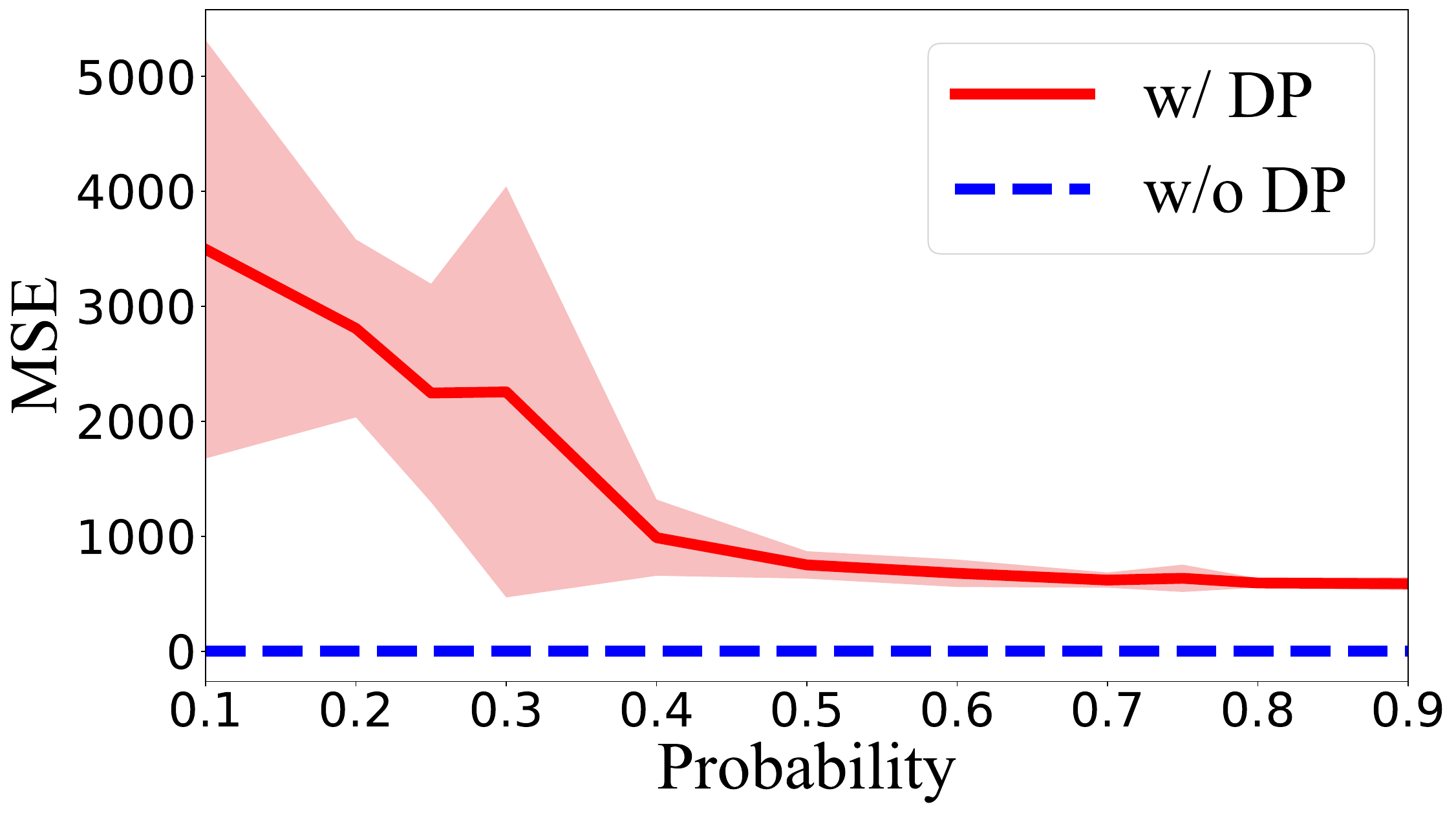}
}

\subfigure[Adult]{   
\includegraphics[width=0.45\textwidth]{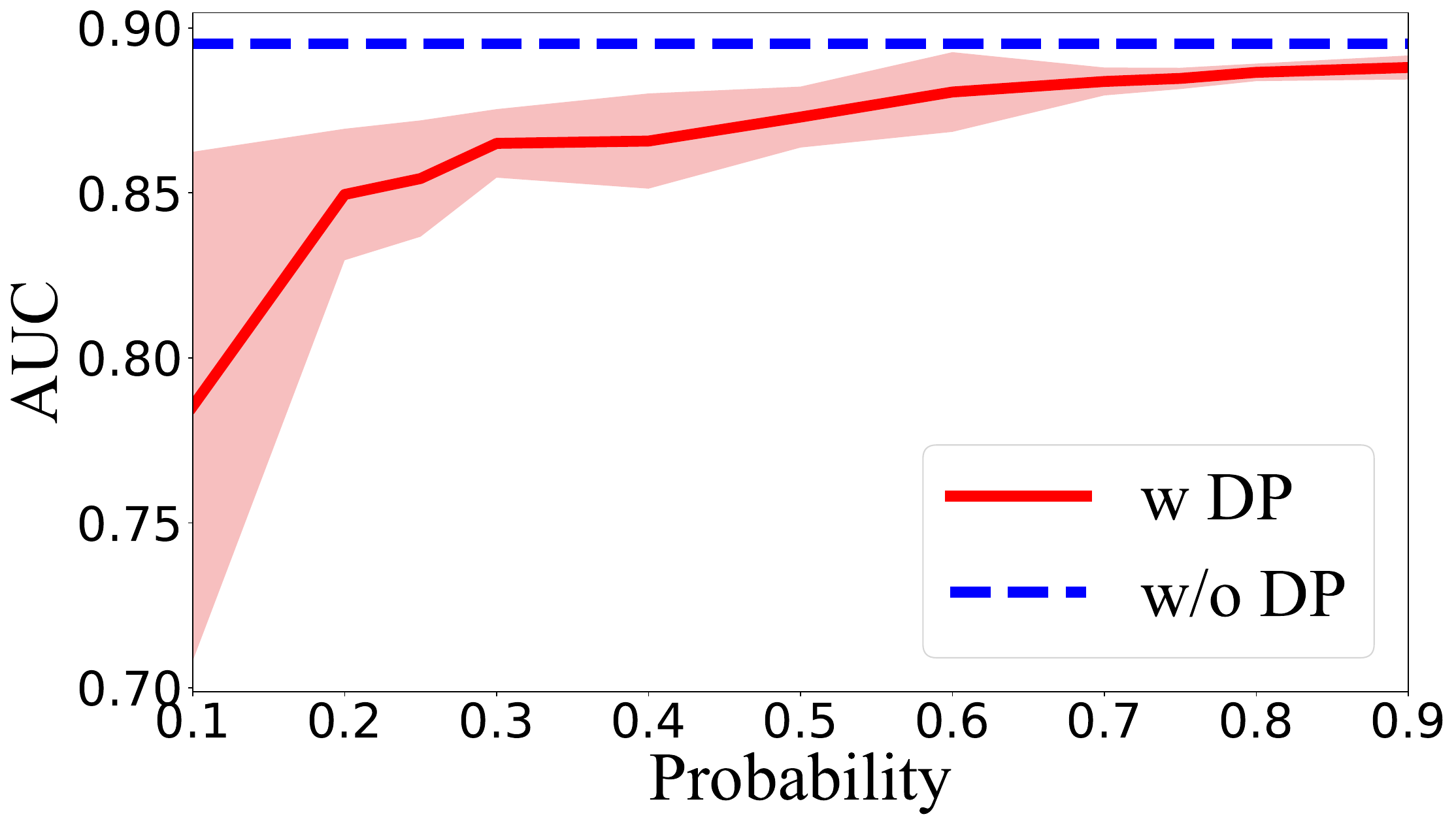}  
}
\subfigure[Credit]{ 
\includegraphics[width=0.45\textwidth]{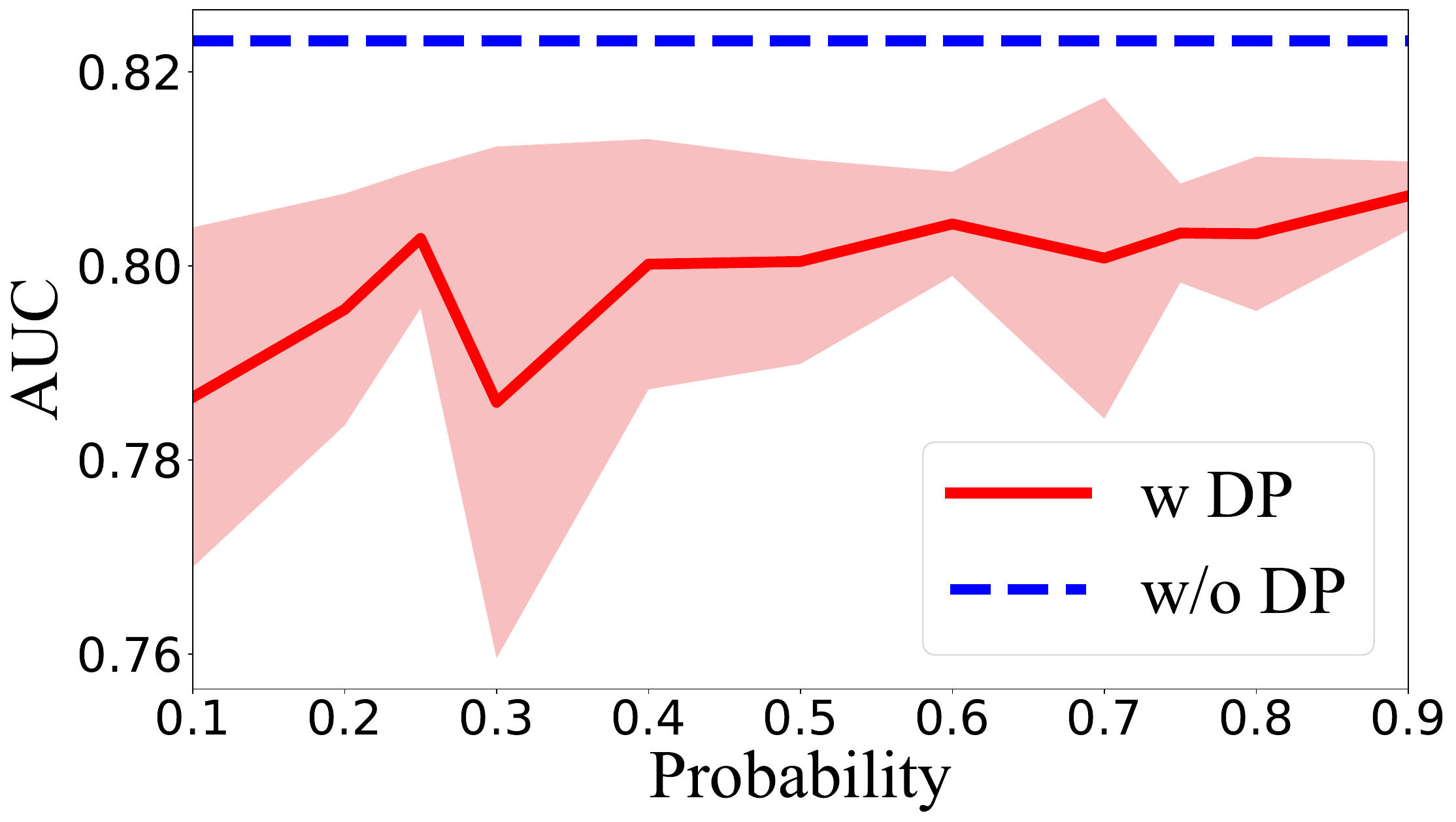}
}
\caption{Model utility w.r.t. different privacy protection strengths when applying FederBoost.\label{fig: diff epsilon for DP} }     
\end{figure*}

\begin{figure*}[t]
\centering  

\subfigure[Abalone]{   
\includegraphics[width=0.45\textwidth]{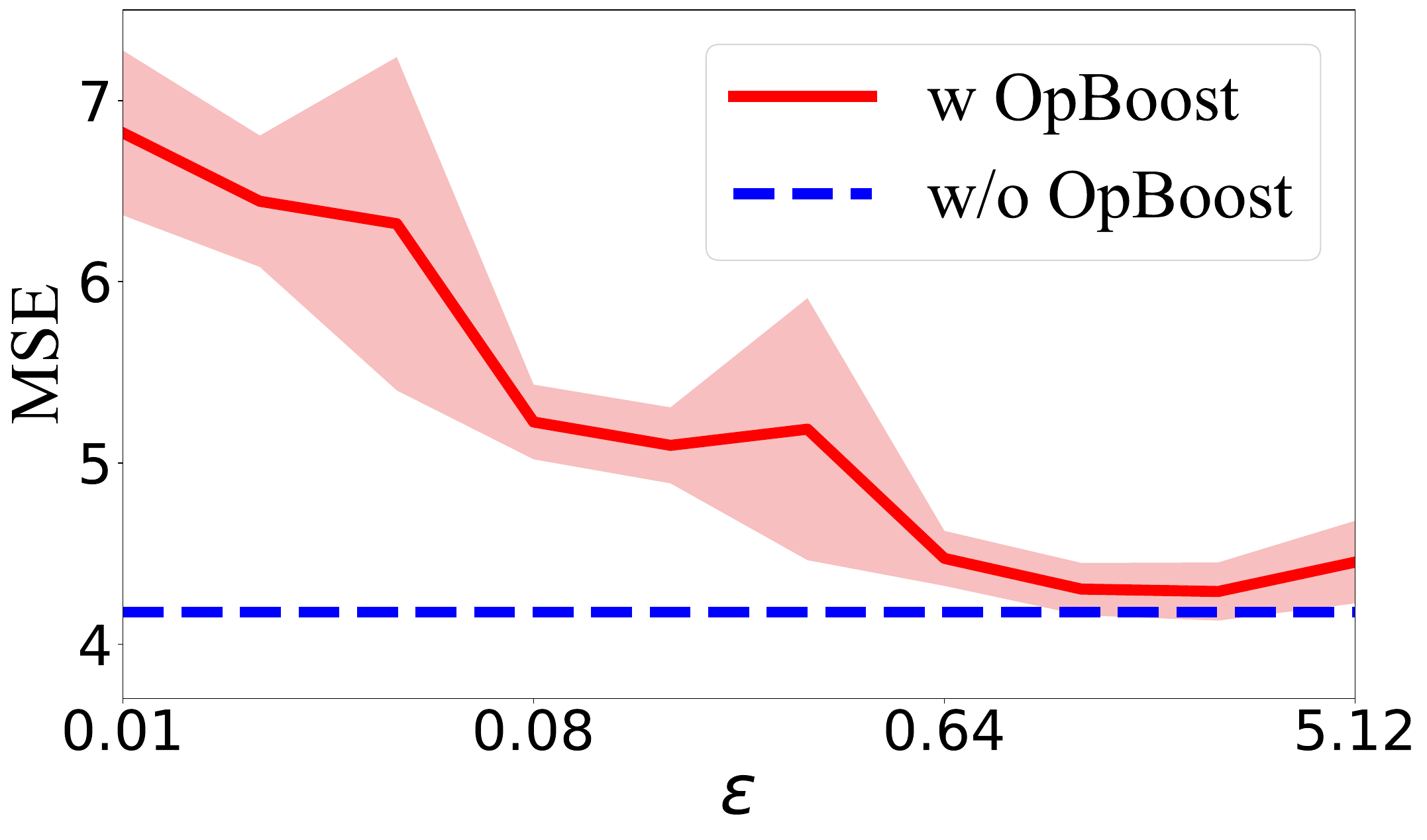} 
}
\subfigure[Blog]{ 
\includegraphics[width=0.45\textwidth]{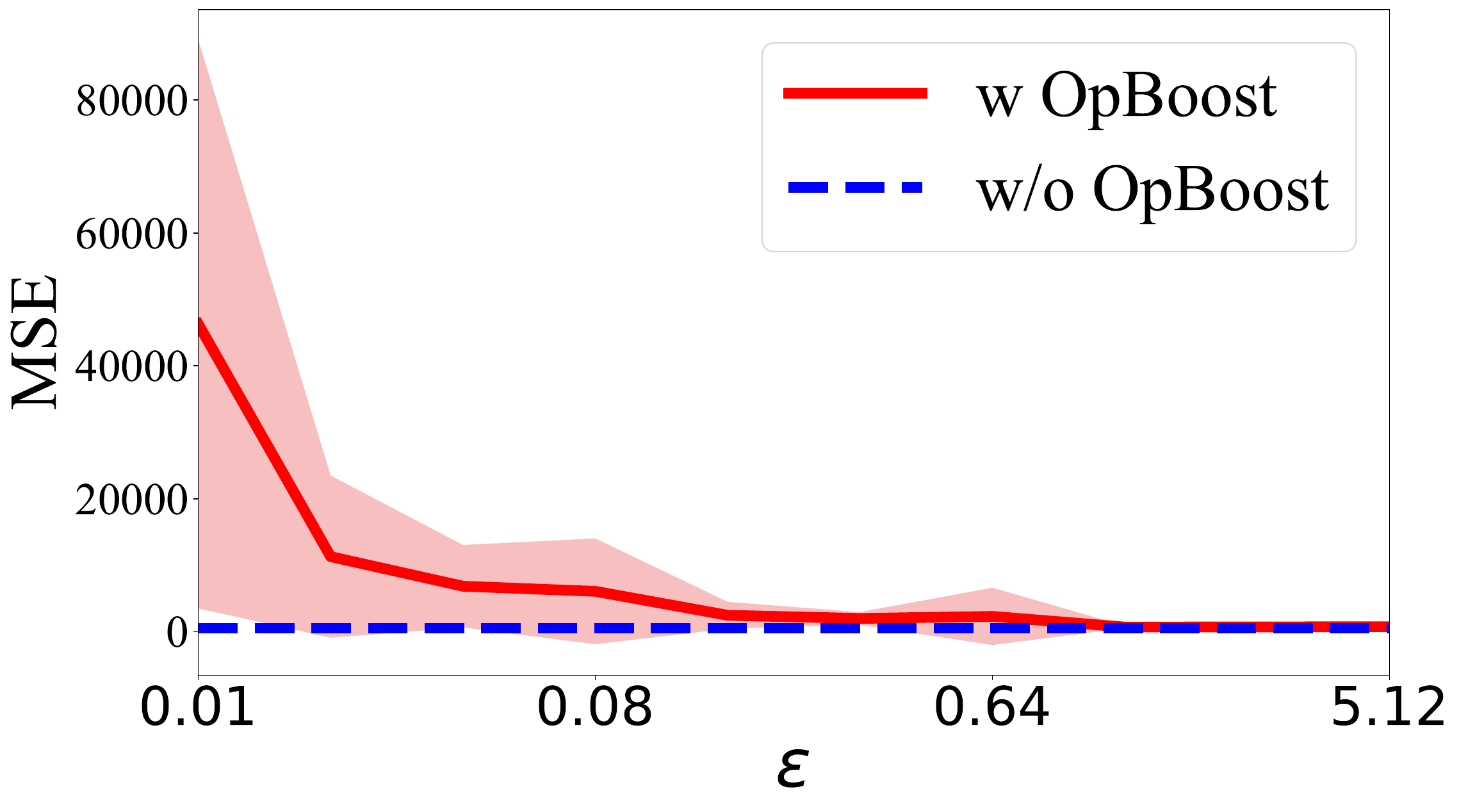}
}
\subfigure[Adult]{   
\includegraphics[width=0.45\textwidth]{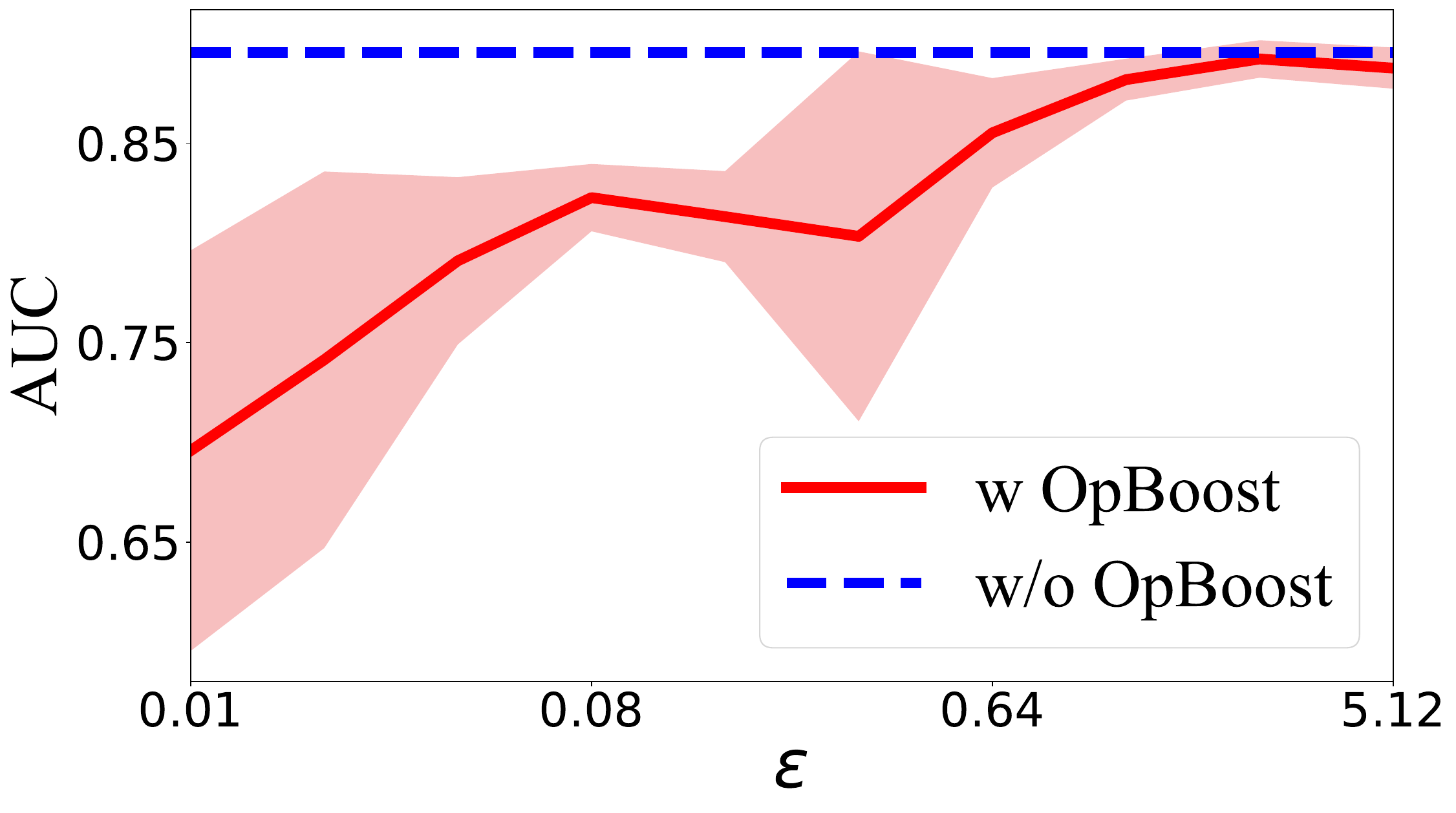}  
}
\subfigure[Credit]{ 
\includegraphics[width=0.45\textwidth]{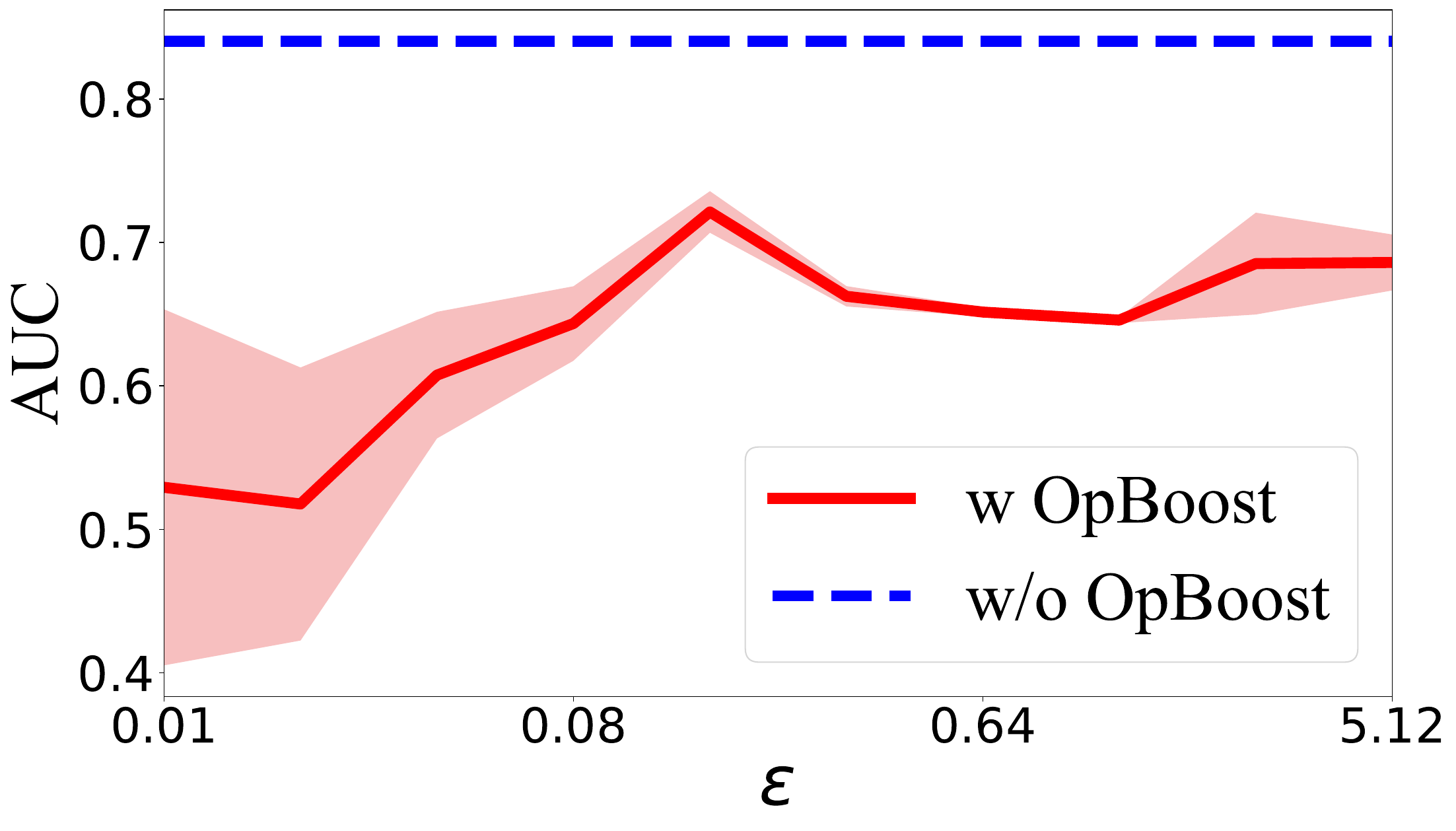}
}
\caption{Model utility w.r.t. different privacy protection strengths when applying OpBoost. \label{fig: diff epsilon for OpBoost}}     
\end{figure*}

In feature-gathering TBMs, in order to prevent privacy leakage, the data parties tend to disrupt ordinal numbers generated based on their private feature. Thus, there exists a trade-off between privacy protection strength and the utility of the model learned based on such desensitized but noisy data. More discussions can be found in Section~\ref{subset:feature_basd_proetction}, and here we provide empirical observation for a better understanding of such a trade-off.
We train XGBoost equipped with the privacy protection mechanisms proposed by FederBoost and OpBoost, and show the experimental results in Figure~\ref{fig: diff epsilon for DP} and Figure~\ref{fig: diff epsilon for OpBoost}, respectively.

As shown at the x-axis in Figure~\ref{fig: diff epsilon for DP}, we vary the probabilities of a sample staying in the correct bucket. From the figure, we can observe that as the probability increases, which implies the intermediate results provided by the data parties are more precise, the model performance becomes better and more stable (\ie fewer variances), and gradually approaches the results achieved by the model learned without adding DP noise.
Similarly, we adjust the privacy protection strength of OpBoost by changing the values of $\epsilon$, which controls the probability of mapping a sample from the $a$-th bucket to $b$-th bucket following $\frac{e^{-|a-b|\cdot\epsilon/2}}{\sum_{j=1}^{50}e^{-|a-j|\cdot\epsilon/2}}$, where $a, b\in [1,50]$, and show the results in Figure~\ref{fig: diff epsilon for OpBoost}. These results further confirm the trade-off between model utility and privacy protection strength when training feature-gathering TBMs.

Note that for label-scattering TBMs, the adopted privacy protection mechanisms might not impact the model utility but bring some efficiency issues, as discussed in the next part.

\subsubsection{Privacy protection strength versus computation and communication cost}
In order to avoid privacy leakage, label-scattering TBMs propose to apply homomorphic encryption and secret sharing techniques, which might bring additional computation and communication costs.
The reason is that the encryption/decryption and framing process needs lots of computation resources, and the encrypted and framed information is always lengthy and costs more computation/communication resources than raw information.

\begin{figure*}[t]
    \begin{minipage}[t]{0.46\textwidth}
        \centering
        \includegraphics[width=.96\textwidth]{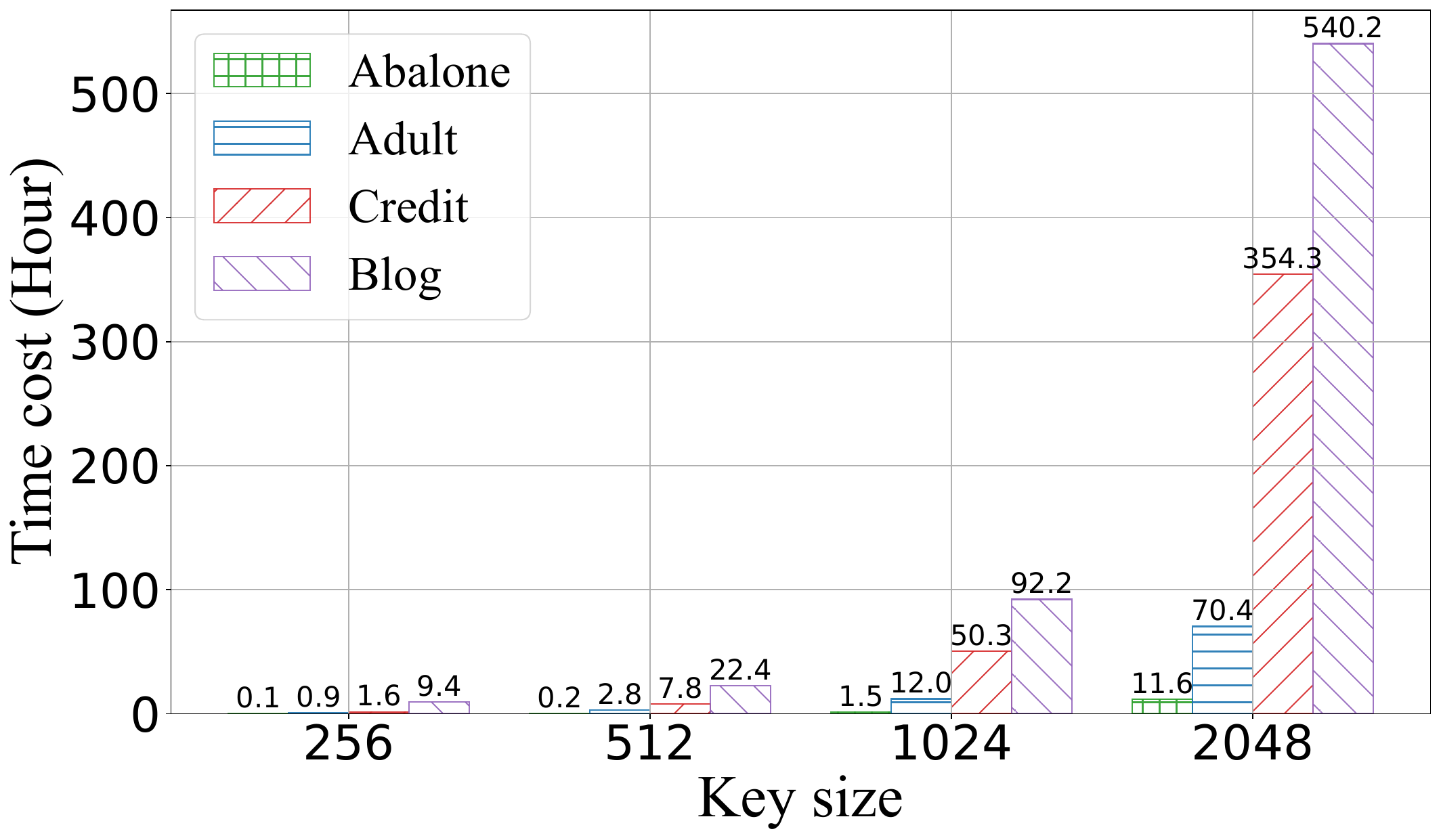}
        \caption{Time cost w.r.t. different key sizes.\label{fig: time consuming for differernt keys}}
    \end{minipage}
     \begin{minipage}[t]{0.46\textwidth}
        \centering
        \includegraphics[width=.96\textwidth]{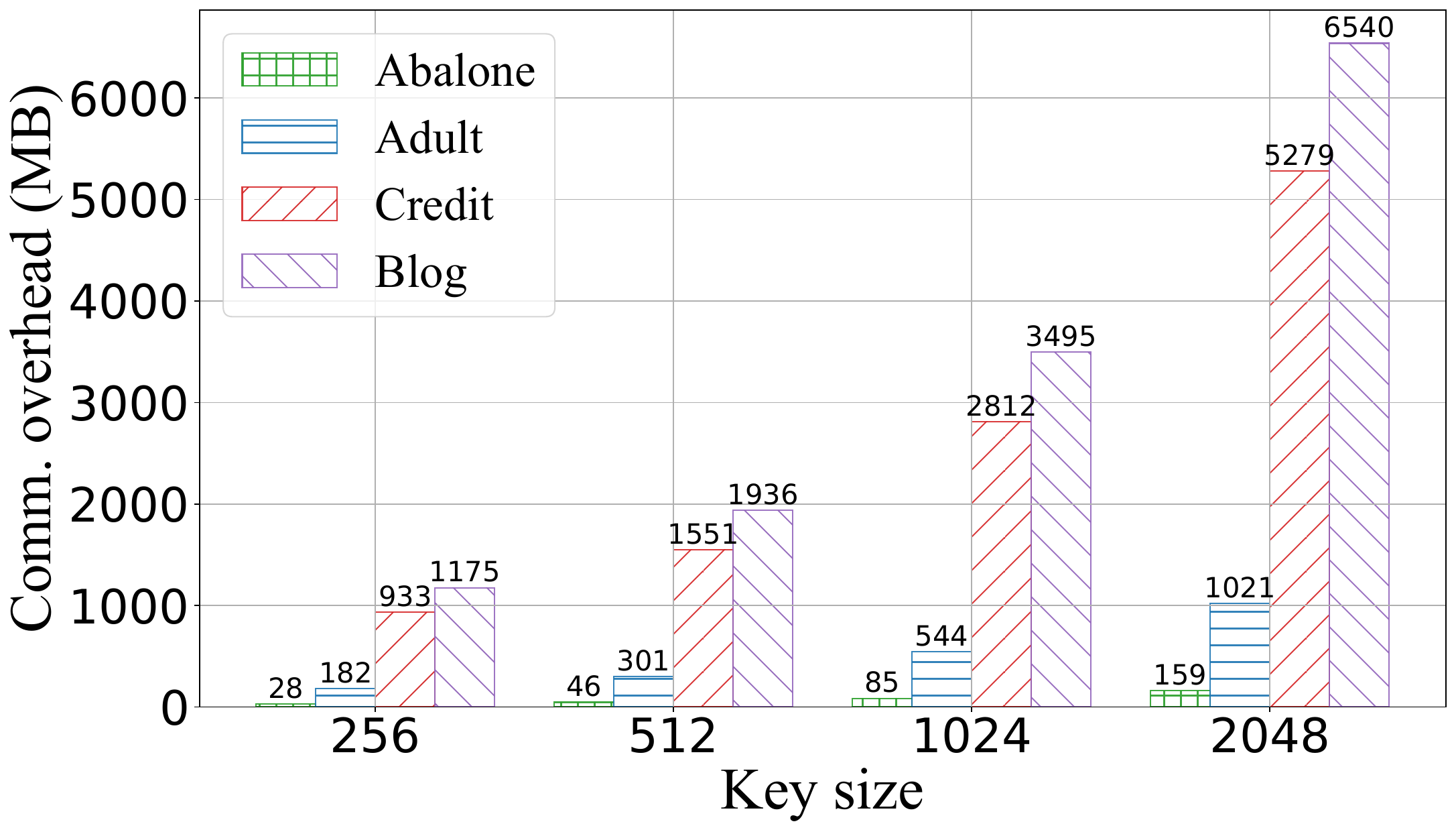}
        \caption{Comm. overhead w.r.t. different key sizes.\label{fig: comm overhead for differernt keys}}
    \end{minipage}
\end{figure*}

In the experiment, we apply Paillier~\cite{paillier1999public} algorithms to encrypt the shared information, and vary the sizes of public and private keys to change the length of encrypted information and thus adjust the privacy protection strength.
We compare the time cost and communication overhead when using different key sizes, and show the results in Figure~\ref{fig: time consuming for differernt keys} and Figure~\ref{fig: comm overhead for differernt keys}, respectively. From these results, we can see that when the key size becomes large, the communication overhead grows linearly, and time consumption (which includes the additional time cost for encrypting, decrypting, sending, receiving, and handling the messages) is growing super linearly. 
Such a phenomenon inspires users to choose a suitable key size when using homomorphic encryption and secret sharing techniques, balancing the privacy protection strength and computation/communication cost.

For feature-gathering TBMs where data parties might inject DP noise into the intermediate results, the communication overhead and time consumption when applying different algorithms are kept at the same level. The slight differences mainly come from the computation and time cost for generating noises and disrupting the ordinal numbers.

\subsubsection{Discussions regarding multiple data parties}

As the number of data parties increases, the utility of the trained TBMs can be influenced by the privacy protection algorithms adopted by each participant. In an extreme case, if each participant either does not employ any privacy protection algorithms or uses performance-preserving algorithms (such as homomorphic encryption), the performance of the TBMs remains consistent regardless of the number of participants. This is primarily because, within VFL, the tree-building process does not fundamentally differ from that in centralized settings, although it distributes the computations among different participants and thus leads to some additional intermediate calculations. On the other hand, when participants utilize differential privacy for enhanced privacy protection, the number of participants affects the model's utility.

An increase in participants results in a proportional rise in communication costs during broadcasting, as we typically utilize a peer-to-peer (P2P) transmission protocol in FL, which implies that each piece of information must be independently transmitted to its designated recipient. Furthermore, the computational costs are closely tied to the number of local features and samples. If the increase in participants leads to a corresponding growth in the number of features and samples (regardless of whether some are redundant), the computational costs would also increase.

It is worth noting that the communication and computation protocols discussed in Section~\ref{sec:method} impose no restrictions on the number of data parties. Therefore, we set the number of data parties to one during the experiments to provide clear insights regarding the trade-offs between model utility, privacy protection, and communication/computation costs.

\subsubsection{Summary}
In a nutshell, both feature-gathering TBMs and label-scattering TBMs can hardly keep model utility, privacy protection strength, and communication/computation cost at the perfect level at the same time.
For example, feature-gathering TBMs are more efficient in achieving an acceptable model utility while exposing desensitized plaintext results, and label-scattering TBMs provide high-level protection strength but need more communication and computation resources.
When applying TBMs in the VFL scenario, researchers and developers are encouraged to propose more advanced algorithms to achieve a better trade-off, and balance model utility, privacy protection strength, and communication/computation cost according to their downstream applications.

\section{Applications \& Directions}
\label{Sec:applications}

In this section, we provide several real-world applications for a better understanding of the advantages of tree-based models in VFL.
Besides, we outline the opportunities and future directions for TBMs in VFL.

\subsection{Real-world Applications}
\label{subsec:applications}

\subsubsection{Finance}
Finance is crucial for social development, encompassing key areas such as risk management and financial marketing. 
Risk management involves identifying, evaluating, and mitigating potential risks to an organization’s capital, earnings, and overall operations. Financial marketing focuses on the strategies that financial institutions use to attract, acquire, and retain customers, including branding, advertising, customer relationship management, and digital marketing.

In practice, financial institutions face challenges due to the lack of comprehensive user profiles essential for effective model training, as well as the limited interpretability of predictive results. Training tree-based models within the federated learning (FL) framework can address these issues by enabling secure collaborations that utilize a wealth of data attributes from external sources, such as internet companies. 
Tree-based models, known for their intuitive structure nature, enhance interpretability by allowing users to easily understand the decision-making process behind predictions. This transparency is crucial for financial institutions, as it fosters trust in model outputs.
In this way, institutions can further improve the accuracy of predictive analytics with actionable insights, and, at the same time, ensure the protection of data privacy.

Recently, notable contributions to this field include vertical federated logistic regression~\cite{hardy2017private}, vertical federated linear regression~\cite{yang2019federated}, and vertical federated boosting tree-based models~\cite{cheng2021secureboost, zhao2022sgboost, xie2022efficient}.
These approaches employ Homomorphic Encryption (HE) and Secret Sharing (SS) to protect the privacy of transmitted intermediate results.

Specifically, WeBank and its partner companies successfully complete vertical federated modeling using invoice data\footnote{https://www.fedai.org/cases/}. This enables WeBank and its partners to collaboratively train the model while keeping their respective data private. The parties exchange encrypted intermediate results and retain control over individual model segments. When it comes time to make predictions, the models from both sides are combined to generate the final prediction. The entire training process prioritizes the security of both the data and the models. Ant Group has announced a financial risk control technology solution based on SMPC that allows for the inclusion of more dimensional credit data into a joint model without compromising the confidentiality of the data sources\footnote{https://www.secretflow.org.cn/en/}. This approach helps to construct more accurate big data credit risk control models. The support of this technology solution enhances the necessary collaboration and communication among participants in financial risk control joint projects. It accelerates the transformation from traditional methods to cutting-edge technologies in financial risk management, serves the collaborative supervision of the government and the financial industry, and promotes the development of the financial data market.

\subsubsection{Recommendation \& Advertising}
Recommendation and advertising are representative domains where VFL holds promising applications. Developing a robust recommendation or advertising system often requires the aggregation of extensive user data from various sources, which raises significant privacy and security concerns. VFL addresses this issue by allowing multiple organizations to collaboratively train a powerful recommendation model without the need to share raw user data.

Consider a scenario where multiple e-commerce platforms aim to develop a combined recommendation or advertising engine that leverages their unique datasets. Each platform contains distinct types of user information, such as purchase history, browsing behavior, user ratings, and demographic details. By utilizing VFL, these platforms can collaboratively train a tree-based model that effectively incorporates the complete feature set available across all platforms, thereby enhancing the relevance of recommendations.

Based on SecureBoost~\cite{cheng2021secureboost}, \cite{song2021federated} has effectively improved recommendation accuracy by leveraging data from mobile network operators and healthcare providers within an FL environment. Besides, research~\cite{liu2022vertical} points out that many companies have successfully implemented tree-based models in VFL for recommendation and advertising systems. For example, ByteDance utilizes the Fedlearner to significantly enhance its advertising effectiveness through the development of a tree-based VFL algorithm; Tencent employs vertical federated GBDTs to establish a VFL federation between advertisers and advertising platforms, resulting in improved model accuracy.

\subsubsection{Healthcare}
VFL has emerged as a transformative technology in the healthcare sector, especially given the fragmented nature of healthcare data. In this landscape, different institutions, such as hospitals, laboratories, and clinics, often hold critical yet incomplete pieces of patient information. These entities typically maintain separate records that encompass diverse aspects of a patient's care, including medical history, diagnostic results, and treatment plans.

These organizations can engage in collaborative analysis without compromising sensitive data privacy with VFL, which allows them to collectively train on a more comprehensive dataset, improving the accuracy, robustness, and interpretability of predictive analytics, while strictly adhering to privacy regulations. 
Vertical federated TBMs hold considerable potential for creating decision rules across multiple data sources to predict patient disease risks, enhance the quality of medical services, improve disease prediction, and optimize treatment protocols.
Tree-based models are also adept at handling variable interactions and providing clear classification rules, which assist physicians in developing personalized treatment plans based on the model’s outputs.

Recent studies~\cite{zheng2023privet, lu2023squirrel, jiang2024sigbdt, akhavan2023level} have designed novel vertical federated TBMs and conducted a series of experiments on healthcare datasets, paving the way for a more integrated approach to healthcare and driving improvements in health outcomes on a larger scale.
Specifically, Ant Group's privacy computing platform, SecretFlow\footnote{https://github.com/secretflow/secretflow}, in collaboration with Alibaba Cloud's medical big data management platform (HData v2.0), utilizes its advanced privacy computing technology to overcome data collaboration barriers among medical institutions in DRGs applications. By integrating artificial intelligence and big data technologies without allowing data to leave the domain, the platform facilitates joint statistics and modeling across institutions. In the realm of medical diagnosis classification, SecretFlow enhances predictive models through the sharing of multi-institutional samples, significantly improving the accuracy of predictions compared to models trained on data from a single institution. The platform is also applied in various scenarios, including disease diagnosis, examination recommendations, medication suggestions, rare disease prediction, and quality control rule management. These applications provide robust solutions for medical data governance, comprehensive hospital quality control, medical research, medical insurance risk management, and critical clinical business challenges.

\subsection{Challenges and Future Directions}
\label{Sec:challenges}

\subsubsection{Communication Overhead} 
Addressing communication overhead during training and inference is a critical area for improvement. This is particularly important considering that the number of transmissions between data parties and label parties in TBMs tends to be higher than in other methods. Meanwhile, communication costs would increase with the number of samples for TBMs. As a result, reducing communication overhead might bring significant improvements in cross-silo FL that involve large datasets.

\subsubsection{Feature Abandonment and Crossing} 
In VFL scenarios, participants do not have access to knowledge about the features owned by others, which can lead to potential redundancy and missed opportunities for creating high-order features through feature crossing. While neural networks may implicitly handle these through non-linear layers, this could be more challenging in TBMs. Thus, enhancing model utility by effectively managing feature abandonment and facilitating feature crossing represents a promising direction, focusing on optimizing the use of available features to improve model accuracy and relevance.

\subsubsection{Balance Among Participants} 
Different participants may possess varying numbers of features, and the importance of these features can be different. An interesting research direction can be proposing mechanisms that ensure the trained model does not become overly biased toward participants with a greater number of features. Conversely, designing suitable incentive mechanisms to reward participants who contribute important features is another valuable area for exploration, leveraging the inherent interpretability of TBMs.

\subsubsection{Non-Overlapping Samples}
Training TBMs in VFL typically rely on overlapping samples, which can limit their applicability. Therefore, designing mechanisms that allow the participation of non-overlapping samples, such as data augmentation strategies, represents a valuable research direction to further broaden the application scope of TBMs in VFL.

\section{Conclusions}
\label{Sec:conclusions}
In this paper, we give a comprehensive survey on tree-based models (TBMs) in vertical federated learning (VFL). We categorize TBMs into feature-gathering TBMs and label-scattering TBMs based on differences in communication and computation protocols. We provide a detailed overview of their training processes and inference procedures, and discuss how to protect various types of shared information using techniques such as differential privacy, homomorphic encryption, and secure multi-party computation. We summarize several design principles aimed at promoting federated learning platforms to support diverse tree-based models for both academic research and industrial deployment. Besides, we provide real-world applications to better illustrate the advancements of TBMs in VFL, including in finance, recommendation systems, advertising, and healthcare, while highlighting some opportunities and future directions.

\bibliographystyle{unsrtnat}
\bibliography{fs-tree}

\begin{thebibliography}{117}
\providecommand{\natexlab}[1]{#1}
\providecommand{\url}[1]{\texttt{#1}}
\expandafter\ifx\csname urlstyle\endcsname\relax
  \providecommand{\doi}[1]{doi: #1}\else
  \providecommand{\doi}{doi: \begingroup \urlstyle{rm}\Url}\fi

\bibitem[Hinton and Salakhutdinov(2006)]{hinton2006reducing}
Geoffrey~E Hinton and Ruslan~R Salakhutdinov.
\newblock Reducing the dimensionality of data with neural networks.
\newblock \emph{science}, 313\penalty0 (5786):\penalty0 504--507, 2006.

\bibitem[LeCun et~al.(2015)LeCun, Bengio, and Hinton]{lecun2015deep}
Yann LeCun, Yoshua Bengio, and Geoffrey Hinton.
\newblock Deep learning.
\newblock \emph{nature}, 521\penalty0 (7553):\penalty0 436--444, 2015.

\bibitem[Chen and Guestrin(2016)]{chen2016xgboost}
Tianqi Chen and Carlos Guestrin.
\newblock Xgboost: A scalable tree boosting system.
\newblock In \emph{Proceedings of the 22nd acm sigkdd international conference on knowledge discovery and data mining}, pages 785--794, 2016.

\bibitem[He et~al.(2016)He, Zhang, Ren, and Sun]{he2016identity}
Kaiming He, Xiangyu Zhang, Shaoqing Ren, and Jian Sun.
\newblock Identity mappings in deep residual networks.
\newblock In \emph{Computer Vision--ECCV 2016: 14th European Conference, Amsterdam, The Netherlands, October 11--14, 2016, Proceedings, Part IV 14}, pages 630--645, 2016.

\bibitem[Krizhevsky et~al.(2017)Krizhevsky, Sutskever, and Hinton]{krizhevsky2017imagenet}
Alex Krizhevsky, Ilya Sutskever, and Geoffrey~E Hinton.
\newblock Imagenet classification with deep convolutional neural networks.
\newblock \emph{Communications of the ACM}, 60\penalty0 (6):\penalty0 84--90, 2017.

\bibitem[Srivastava et~al.(2014)Srivastava, Hinton, Krizhevsky, Sutskever, and Salakhutdinov]{srivastava2014dropout}
Nitish Srivastava, Geoffrey Hinton, Alex Krizhevsky, Ilya Sutskever, and Ruslan Salakhutdinov.
\newblock Dropout: a simple way to prevent neural networks from overfitting.
\newblock \emph{The journal of machine learning research}, 15\penalty0 (1):\penalty0 1929--1958, 2014.

\bibitem[Vaswani et~al.(2017)Vaswani, Shazeer, Parmar, Uszkoreit, Jones, Gomez, Kaiser, and Polosukhin]{vaswani2017attention}
Ashish Vaswani, Noam Shazeer, Niki Parmar, Jakob Uszkoreit, Llion Jones, Aidan~N Gomez, {\L}ukasz Kaiser, and Illia Polosukhin.
\newblock Attention is all you need.
\newblock \emph{Advances in neural information processing systems}, 30, 2017.

\bibitem[Jordan and Mitchell(2015)]{machine2015jordan}
M.~I. Jordan and T.~M. Mitchell.
\newblock Machine learning: Trends, perspectives, and prospects.
\newblock \emph{Science}, 349\penalty0 (6245):\penalty0 255--260, 2015.

\bibitem[Li et~al.(2021{\natexlab{a}})Li, Wen, Wu, Hu, Wang, Li, Liu, and He]{survey2021li}
Qinbin Li, Zeyi Wen, Zhaomin Wu, Sixu Hu, Naibo Wang, Yuan Li, Xu~Liu, and Bingsheng He.
\newblock A survey on federated learning systems: Vision, hype and reality for data privacy and protection.
\newblock \emph{IEEE Transactions on Knowledge and Data Engineering}, 2021{\natexlab{a}}.

\bibitem[McMahan et~al.(2017)McMahan, Moore, Ramage, Hampson, and y~Arcas]{mcmahan2017communication}
Brendan McMahan, Eider Moore, Daniel Ramage, Seth Hampson, and Blaise~Aguera y~Arcas.
\newblock Communication-efficient learning of deep networks from decentralized data.
\newblock In \emph{Artificial intelligence and statistics}, pages 1273--1282, 2017.

\bibitem[Kairouz et~al.(2021)Kairouz, McMahan, Avent, Bellet, Bennis, Bhagoji, Bonawitz, Charles, Cormode, Cummings, et~al.]{kairouz2021advances}
Peter Kairouz, H~Brendan McMahan, Brendan Avent, Aur{\'e}lien Bellet, Mehdi Bennis, Arjun~Nitin Bhagoji, Kallista Bonawitz, Zachary Charles, Graham Cormode, Rachel Cummings, et~al.
\newblock Advances and open problems in federated learning.
\newblock \emph{Foundations and Trends{\textregistered} in Machine Learning}, 14\penalty0 (1--2):\penalty0 1--210, 2021.

\bibitem[Yang et~al.(2019{\natexlab{a}})Yang, Liu, Chen, and Tong]{yang2019federated}
Qiang Yang, Yang Liu, Tianjian Chen, and Yongxin Tong.
\newblock Federated machine learning: Concept and applications.
\newblock \emph{ACM Transactions on Intelligent Systems and Technology (TIST)}, 10\penalty0 (2):\penalty0 1--19, 2019{\natexlab{a}}.

\bibitem[Chen et~al.(2021{\natexlab{a}})Chen, Zhou, Wang, Wu, Fang, Tan, Wang, Liu, Wang, and Hong]{chen2021homomorphic}
Chaochao Chen, Jun Zhou, Li~Wang, Xibin Wu, Wenjing Fang, Jin Tan, Lei Wang, Alex~X Liu, Hao Wang, and Cheng Hong.
\newblock When homomorphic encryption marries secret sharing: Secure large-scale sparse logistic regression and applications in risk control.
\newblock In \emph{Proceedings of the 27th ACM SIGKDD Conference on Knowledge Discovery \& Data Mining}, pages 2652--2662, 2021{\natexlab{a}}.

\bibitem[Long et~al.(2020)Long, Tan, Jiang, and Zhang]{long2020federated}
Guodong Long, Yue Tan, Jing Jiang, and Chengqi Zhang.
\newblock Federated learning for open banking.
\newblock In \emph{Federated learning}, pages 240--254. 2020.

\bibitem[Wang(2019)]{wang2019interpret}
Guan Wang.
\newblock Interpret federated learning with shapley values.
\newblock \emph{arXiv preprint arXiv:1905.04519}, 2019.

\bibitem[Cheng et~al.(2020)Cheng, Liu, Chen, and Yang]{cheng2020federated}
Yong Cheng, Yang Liu, Tianjian Chen, and Qiang Yang.
\newblock Federated learning for privacy-preserving ai.
\newblock \emph{Communications of the ACM}, 63\penalty0 (12):\penalty0 33--36, 2020.

\bibitem[Ammad-Ud-Din et~al.(2019)Ammad-Ud-Din, Ivannikova, Khan, Oyomno, Fu, Tan, and Flanagan]{ammad2019federated}
Muhammad Ammad-Ud-Din, Elena Ivannikova, Suleiman~A Khan, Were Oyomno, Qiang Fu, Kuan~Eeik Tan, and Adrian Flanagan.
\newblock Federated collaborative filtering for privacy-preserving personalized recommendation system.
\newblock \emph{arXiv preprint arXiv:1901.09888}, 2019.

\bibitem[Zhang and Jiang(2021)]{zhang2021vertical}
JianFei Zhang and YuChen Jiang.
\newblock A vertical federation recommendation method based on clustering and latent factor model.
\newblock In \emph{2021 International Conference on Electronic Information Engineering and Computer Science (EIECS)}, pages 362--366, 2021.

\bibitem[Cui et~al.(2021)Cui, Chen, Lyu, Yang, and Li]{cui2021exploiting}
Jinming Cui, Chaochao Chen, Lingjuan Lyu, Carl Yang, and Wang Li.
\newblock Exploiting data sparsity in secure cross-platform social recommendation.
\newblock \emph{Advances in Neural Information Processing Systems}, 34:\penalty0 10524--10534, 2021.

\bibitem[Shmueli and Tassa(2017)]{shmueli2017secure}
Erez Shmueli and Tamir Tassa.
\newblock Secure multi-party protocols for item-based collaborative filtering.
\newblock In \emph{Proceedings of the eleventh ACM conference on recommender systems}, pages 89--97, 2017.

\bibitem[Leo et~al.(2019)Leo, Sharma, and Maddulety]{leo2019machine}
Martin Leo, Suneel Sharma, and Koilakuntla Maddulety.
\newblock Machine learning in banking risk management: A literature review.
\newblock \emph{Risks}, 7\penalty0 (1):\penalty0 29, 2019.

\bibitem[Zheng et~al.(2022)Zheng, Zhou, Sun, Wang, Liu, and Li]{zheng2022applications}
Zhaohua Zheng, Yize Zhou, Yilong Sun, Zhang Wang, Boyi Liu, and Keqiu Li.
\newblock Applications of federated learning in smart cities: recent advances, taxonomy, and open challenges.
\newblock \emph{Connection Science}, 34\penalty0 (1):\penalty0 1--28, 2022.

\bibitem[Jiang et~al.(2020)Jiang, Kantarci, Oktug, and Soyata]{jiang2020federated}
Ji~Chu Jiang, Burak Kantarci, Sema Oktug, and Tolga Soyata.
\newblock Federated learning in smart city sensing: Challenges and opportunities.
\newblock \emph{Sensors}, 20\penalty0 (21):\penalty0 6230, 2020.

\bibitem[Ramu et~al.(2022)Ramu, Boopalan, Pham, Maddikunta, Huynh-The, Alazab, Nguyen, and Gadekallu]{ramu2022federated}
Swarna~Priya Ramu, Parimala Boopalan, Quoc-Viet Pham, Praveen Kumar~Reddy Maddikunta, Thien Huynh-The, Mamoun Alazab, Thanh~Thi Nguyen, and Thippa~Reddy Gadekallu.
\newblock Federated learning enabled digital twins for smart cities: Concepts, recent advances, and future directions.
\newblock \emph{Sustainable Cities and Society}, 79:\penalty0 103663, 2022.

\bibitem[Chen et~al.(2020{\natexlab{a}})Chen, Jin, Sun, and Yin]{chen2020vafl}
Tianyi Chen, Xiao Jin, Yuejiao Sun, and Wotao Yin.
\newblock Vafl: a method of vertical asynchronous federated learning.
\newblock \emph{arXiv preprint arXiv:2007.06081}, 2020{\natexlab{a}}.

\bibitem[Teimoori et~al.(2022)Teimoori, Yassine, and Hossain]{teimoori2022secure}
Zeinab Teimoori, Abdulsalam Yassine, and M~Shamim Hossain.
\newblock A secure cloudlet-based charging station recommendation for electric vehicles empowered by federated learning.
\newblock \emph{IEEE Transactions on Industrial Informatics}, 18\penalty0 (9):\penalty0 6464--6473, 2022.

\bibitem[Chen et~al.(2020{\natexlab{b}})Chen, Zhou, Zheng, Wu, Lyu, Wu, Wu, Liu, Wang, and Zheng]{chen2020vertically}
Chaochao Chen, Jun Zhou, Longfei Zheng, Huiwen Wu, Lingjuan Lyu, Jia Wu, Bingzhe Wu, Ziqi Liu, Li~Wang, and Xiaolin Zheng.
\newblock Vertically federated graph neural network for privacy-preserving node classification.
\newblock \emph{arXiv preprint arXiv:2005.11903}, 2020{\natexlab{b}}.

\bibitem[He et~al.(2020{\natexlab{a}})He, Annavaram, and Avestimehr]{he2020group}
Chaoyang He, Murali Annavaram, and Salman Avestimehr.
\newblock Group knowledge transfer: Federated learning of large cnns at the edge.
\newblock \emph{Advances in Neural Information Processing Systems}, 33:\penalty0 14068--14080, 2020{\natexlab{a}}.

\bibitem[Gupta and Raskar(2018)]{gupta2018distributed}
Otkrist Gupta and Ramesh Raskar.
\newblock Distributed learning of deep neural network over multiple agents.
\newblock \emph{Journal of Network and Computer Applications}, 116:\penalty0 1--8, 2018.

\bibitem[Dang et~al.(2020)Dang, Gu, and Huang]{dang2020large}
Zhiyuan Dang, Bin Gu, and Heng Huang.
\newblock Large-scale kernel method for vertical federated learning.
\newblock In \emph{Federated Learning}, pages 66--80. 2020.

\bibitem[Gu et~al.(2020)Gu, Dang, Li, and Huang]{gu2020federated}
Bin Gu, Zhiyuan Dang, Xiang Li, and Heng Huang.
\newblock Federated doubly stochastic kernel learning for vertically partitioned data.
\newblock In \emph{Proceedings of the 26th ACM SIGKDD international conference on knowledge discovery \& data mining}, pages 2483--2493, 2020.

\bibitem[Yang et~al.(2019{\natexlab{b}})Yang, Ren, Zhou, and Liu]{yang2019parallel}
Shengwen Yang, Bing Ren, Xuhui Zhou, and Liping Liu.
\newblock Parallel distributed logistic regression for vertical federated learning without third-party coordinator.
\newblock \emph{arXiv preprint arXiv:1911.09824}, 2019{\natexlab{b}}.

\bibitem[Bonawitz et~al.(2017)Bonawitz, Ivanov, Kreuter, Marcedone, McMahan, Patel, Ramage, Segal, and Seth]{bonawitz2017practical}
Keith Bonawitz, Vladimir Ivanov, Ben Kreuter, Antonio Marcedone, H~Brendan McMahan, Sarvar Patel, Daniel Ramage, Aaron Segal, and Karn Seth.
\newblock Practical secure aggregation for privacy-preserving machine learning.
\newblock In \emph{proceedings of the 2017 ACM SIGSAC Conference on Computer and Communications Security}, pages 1175--1191, 2017.

\bibitem[Romanini et~al.(2021)Romanini, Hall, Papadopoulos, Titcombe, Ismail, Cebere, Sandmann, Roehm, and Hoeh]{romanini2021pyvertical}
Daniele Romanini, Adam~James Hall, Pavlos Papadopoulos, Tom Titcombe, Abbas Ismail, Tudor Cebere, Robert Sandmann, Robin Roehm, and Michael~A Hoeh.
\newblock Pyvertical: A vertical federated learning framework for multi-headed splitnn.
\newblock \emph{arXiv preprint arXiv:2104.00489}, 2021.

\bibitem[Liu et~al.(2020)Liu, Liu, Liu, Liang, Meng, Zhang, and Zheng]{liu2020federated}
Yang Liu, Yingting Liu, Zhijie Liu, Yuxuan Liang, Chuishi Meng, Junbo Zhang, and Yu~Zheng.
\newblock Federated forest.
\newblock \emph{IEEE Transactions on Big Data}, 2020.

\bibitem[Cheng et~al.(2021)Cheng, Fan, Jin, Liu, Chen, Papadopoulos, and Yang]{cheng2021secureboost}
Kewei Cheng, Tao Fan, Yilun Jin, Yang Liu, Tianjian Chen, Dimitrios Papadopoulos, and Qiang Yang.
\newblock Secureboost: A lossless federated learning framework.
\newblock \emph{IEEE Intelligent Systems}, 36\penalty0 (6):\penalty0 87--98, 2021.

\bibitem[Fang et~al.(2021)Fang, Zhao, Tan, Chen, Yu, Wang, Wang, Zhou, and Zhang]{fang2021large}
Wenjing Fang, Derun Zhao, Jin Tan, Chaochao Chen, Chaofan Yu, Li~Wang, Lei Wang, Jun Zhou, and Benyu Zhang.
\newblock Large-scale secure xgb for vertical federated learning.
\newblock In \emph{Proceedings of the 30th ACM International Conference on Information \& Knowledge Management}, pages 443--452, 2021.

\bibitem[Tian et~al.(2024)Tian, Zhang, Hou, Lyu, Zhang, Liu, and Ren]{tian2020federboost}
Zhihua Tian, Rui Zhang, Xiaoyang Hou, Lingjuan Lyu, Tianyi Zhang, Jian Liu, and Kui Ren.
\newblock Federboost: Private federated learning for gbdt.
\newblock \emph{IEEE Transactions on Dependable and Secure Computing}, 21\penalty0 (3):\penalty0 1274--1285, 2024.

\bibitem[Li et~al.(2022)Li, Hu, Liu, Feng, Peng, Hong, Ren, and Qin]{li2022opboost}
Xiaochen Li, Yuke Hu, Weiran Liu, Hanwen Feng, Li~Peng, Yuan Hong, Kui Ren, and Zhan Qin.
\newblock Opboost: a vertical federated tree boosting framework based on order-preserving desensitization.
\newblock \emph{Proceedings of the VLDB Endowment}, 16\penalty0 (2):\penalty0 202--215, 2022.

\bibitem[Chen et~al.(2021{\natexlab{b}})Chen, Zhou, Guan, Yang, Fao, Wang, and Wang]{chen2021fed}
Xiaolin Chen, Shuai Zhou, Bei Guan, Kai Yang, Hao Fao, Hu~Wang, and Yongji Wang.
\newblock Fed-eini: An efficient and interpretable inference framework for decision tree ensembles in vertical federated learning.
\newblock In \emph{2021 IEEE International Conference on Big Data (Big Data)}, pages 1242--1248, 2021{\natexlab{b}}.

\bibitem[Song et~al.(2021)Song, Xie, Zhang, Liang, Ye, Yang, and Ouyang]{song2021federated}
Yong Song, Yuchen Xie, Hongwei Zhang, Yuxin Liang, Xiaozhou Ye, Aidong Yang, and Ye~Ouyang.
\newblock Federated learning application on telecommunication-joint healthcare recommendation.
\newblock In \emph{2021 IEEE 21st International Conference on Communication Technology (ICCT)}, pages 1443--1448, 2021.

\bibitem[Jin et~al.(2022)Jin, Wang, Teo, Zhang, Chan, Hou, and Aung]{jin2022towards}
Chao Jin, Jun Wang, Sin~G Teo, Le~Zhang, C~Chan, Qibin Hou, and Khin Mi~Mi Aung.
\newblock Towards end-to-end secure and efficient federated learning for xgboost.
\newblock In \emph{Proceedings of the AAAI International Workshop on Trustable, Verifiable and Auditable Federated Learning}, 2022.

\bibitem[Zheng et~al.(2023)Zheng, Xu, Wang, Gao, and Hua]{zheng2023privet}
Yifeng Zheng, Shuangqing Xu, Songlei Wang, Yansong Gao, and Zhongyun Hua.
\newblock Privet: A privacy-preserving vertical federated learning service for gradient boosted decision tables.
\newblock \emph{IEEE Transactions on Services Computing}, 16\penalty0 (5):\penalty0 3604--3620, 2023.

\bibitem[Lu et~al.(2023)Lu, Huang, Zhang, Wang, and Hong]{lu2023squirrel}
Wen-jie Lu, Zhicong Huang, Qizhi Zhang, Yuchen Wang, and Cheng Hong.
\newblock Squirrel: A scalable secure $\{$Two-Party$\}$ computation framework for training gradient boosting decision tree.
\newblock In \emph{32nd USENIX Security Symposium (USENIX Security 23)}, pages 6435--6451, 2023.

\bibitem[Jiang et~al.(2024)Jiang, Mei, Dai, and Li]{jiang2024sigbdt}
Yufan Jiang, Fei Mei, Tianxiang Dai, and Yong Li.
\newblock Sigbdt: Large-scale gradient boosting decision tree training via function secret sharing.
\newblock In \emph{Proceedings of the 19th ACM Asia Conference on Computer and Communications Security}, pages 274--288, 2024.

\bibitem[Akhavan~Mahdavi et~al.(2023)Akhavan~Mahdavi, Ni, Linkov, and Kerschbaum]{akhavan2023level}
Rasoul Akhavan~Mahdavi, Haoyan Ni, Dimitry Linkov, and Florian Kerschbaum.
\newblock Level up: Private non-interactive decision tree evaluation using levelled homomorphic encryption.
\newblock In \emph{Proceedings of the 2023 ACM SIGSAC Conference on Computer and Communications Security}, pages 2945--2958, 2023.

\bibitem[Xu et~al.(2024)Xu, Zhu, Zheng, Wang, Zhao, Liu, and Li]{xu2024elxgb}
Wei Xu, Hui Zhu, Yandong Zheng, Fengwei Wang, Jiaqi Zhao, Zhe Liu, and Hui Li.
\newblock Elxgb: An efficient and privacy-preserving xgboost for vertical federated learning.
\newblock \emph{IEEE Transactions on Services Computing}, 2024.

\bibitem[Chen et~al.(2022)Chen, Li, Wang, Hao, Xu, and Zhang]{chen2022privdt}
Hanxiao Chen, Hongwei Li, Yingzhe Wang, Meng Hao, Guowen Xu, and Tianwei Zhang.
\newblock Privdt: An efficient two-party cryptographic framework for vertical decision trees.
\newblock \emph{IEEE Transactions on Information Forensics and Security}, 18:\penalty0 1006--1021, 2022.

\bibitem[Xia et~al.(2022)Xia, Zheng, Li, Tang, and Zhang]{xia2022privacy}
Liqiao Xia, Pai Zheng, Jinjie Li, Wangchujun Tang, and Xiangying Zhang.
\newblock Privacy-preserving gradient boosting tree: Vertical federated learning for collaborative bearing fault diagnosis.
\newblock \emph{IET Collaborative Intelligent Manufacturing}, 4\penalty0 (3):\penalty0 208--219, 2022.

\bibitem[Grinsztajn et~al.(2022)Grinsztajn, Oyallon, and Varoquaux]{grinsztajn2022tree}
Leo Grinsztajn, Edouard Oyallon, and Gael Varoquaux.
\newblock Why do tree-based models still outperform deep learning on typical tabular data?
\newblock In \emph{Thirty-sixth Conference on Neural Information Processing Systems Datasets and Benchmarks Track}, 2022.

\bibitem[Dwork(2008)]{dwork2008differential}
Cynthia Dwork.
\newblock Differential privacy: A survey of results.
\newblock In \emph{International conference on theory and applications of models of computation}, pages 1--19, 2008.

\bibitem[Abadi et~al.(2016)Abadi, Chu, Goodfellow, McMahan, Mironov, Talwar, and Zhang]{abadi2016deep}
Martin Abadi, Andy Chu, Ian Goodfellow, H~Brendan McMahan, Ilya Mironov, Kunal Talwar, and Li~Zhang.
\newblock Deep learning with differential privacy.
\newblock In \emph{Proceedings of the 2016 ACM SIGSAC conference on computer and communications security}, pages 308--318, 2016.

\bibitem[Dwork et~al.(2014)Dwork, Roth, et~al.]{dwork2014algorithmic}
Cynthia Dwork, Aaron Roth, et~al.
\newblock The algorithmic foundations of differential privacy.
\newblock \emph{Foundations and Trends{\textregistered} in Theoretical Computer Science}, 9\penalty0 (3--4):\penalty0 211--407, 2014.

\bibitem[Paillier(1999)]{paillier1999public}
Pascal Paillier.
\newblock Public-key cryptosystems based on composite degree residuosity classes.
\newblock In \emph{Advances in Cryptology—EUROCRYPT’99: International Conference on the Theory and Application of Cryptographic Techniques Prague, Czech Republic, May 2--6, 1999 Proceedings 18}, pages 223--238, 1999.

\bibitem[Yi et~al.(2014)Yi, Paulet, Bertino, Yi, Paulet, and Bertino]{yi2014homomorphic}
Xun Yi, Russell Paulet, Elisa Bertino, Xun Yi, Russell Paulet, and Elisa Bertino.
\newblock \emph{Homomorphic encryption}.
\newblock 2014.

\bibitem[Fontaine and Galand(2007)]{fontaine2007survey}
Caroline Fontaine and Fabien Galand.
\newblock A survey of homomorphic encryption for nonspecialists.
\newblock \emph{EURASIP Journal on Information Security}, 2007:\penalty0 1--10, 2007.

\bibitem[Acar et~al.(2018)Acar, Aksu, Uluagac, and Conti]{acar2018survey}
Abbas Acar, Hidayet Aksu, A~Selcuk Uluagac, and Mauro Conti.
\newblock A survey on homomorphic encryption schemes: Theory and implementation.
\newblock \emph{ACM Computing Surveys (Csur)}, 51\penalty0 (4):\penalty0 1--35, 2018.

\bibitem[Naehrig et~al.(2011)Naehrig, Lauter, and Vaikuntanathan]{naehrig2011can}
Michael Naehrig, Kristin Lauter, and Vinod Vaikuntanathan.
\newblock Can homomorphic encryption be practical?
\newblock In \emph{Proceedings of the 3rd ACM workshop on Cloud computing security workshop}, pages 113--124, 2011.

\bibitem[Gentry(2009)]{gentry2009fully}
Craig Gentry.
\newblock Fully homomorphic encryption using ideal lattices.
\newblock In \emph{Proceedings of the forty-first annual ACM symposium on Theory of computing}, pages 169--178, 2009.

\bibitem[Yao(1982)]{yao1982protocols}
Andrew~C Yao.
\newblock Protocols for secure computations.
\newblock In \emph{23rd annual symposium on foundations of computer science (sfcs 1982)}, pages 160--164, 1982.

\bibitem[Goldreich(1998)]{goldreich1998secure}
Oded Goldreich.
\newblock Secure multi-party computation.
\newblock \emph{Manuscript. Preliminary version}, 78\penalty0 (110), 1998.

\bibitem[Du and Atallah(2001)]{du2001secure}
Wenliang Du and Mikhail~J Atallah.
\newblock Secure multi-party computation problems and their applications: a review and open problems.
\newblock In \emph{Proceedings of the 2001 workshop on New security paradigms}, pages 13--22, 2001.

\bibitem[Zhao et~al.(2019)Zhao, Zhao, Zhao, Chen, Gao, Li, and Tan]{zhao2019secure}
Chuan Zhao, Shengnan Zhao, Minghao Zhao, Zhenxiang Chen, Chong-Zhi Gao, Hongwei Li, and Yu-an Tan.
\newblock Secure multi-party computation: theory, practice and applications.
\newblock \emph{Information Sciences}, 476:\penalty0 357--372, 2019.

\bibitem[Evans et~al.(2018)Evans, Kolesnikov, Rosulek, et~al.]{evans2018pragmatic}
David Evans, Vladimir Kolesnikov, Mike Rosulek, et~al.
\newblock A pragmatic introduction to secure multi-party computation.
\newblock \emph{Foundations and Trends{\textregistered} in Privacy and Security}, 2\penalty0 (2-3):\penalty0 70--246, 2018.

\bibitem[Ben-David et~al.(2008)Ben-David, Nisan, and Pinkas]{ben2008fairplaymp}
Assaf Ben-David, Noam Nisan, and Benny Pinkas.
\newblock Fairplaymp: a system for secure multi-party computation.
\newblock In \emph{Proceedings of the 15th ACM conference on Computer and communications security}, pages 257--266, 2008.

\bibitem[Li et~al.(2020)Li, Sahu, Talwalkar, and Smith]{li2020federated}
Tian Li, Anit~Kumar Sahu, Ameet Talwalkar, and Virginia Smith.
\newblock Federated learning: Challenges, methods, and future directions.
\newblock \emph{IEEE signal processing magazine}, 37\penalty0 (3):\penalty0 50--60, 2020.

\bibitem[Aledhari et~al.(2020)Aledhari, Razzak, Parizi, and Saeed]{aledhari2020federated}
Mohammed Aledhari, Rehma Razzak, Reza~M Parizi, and Fahad Saeed.
\newblock Federated learning: A survey on enabling technologies, protocols, and applications.
\newblock \emph{IEEE Access}, 8:\penalty0 140699--140725, 2020.

\bibitem[Li et~al.(2021{\natexlab{b}})Li, Wen, Wu, Hu, Wang, Li, Liu, and He]{li2021survey}
Qinbin Li, Zeyi Wen, Zhaomin Wu, Sixu Hu, Naibo Wang, Yuan Li, Xu~Liu, and Bingsheng He.
\newblock A survey on federated learning systems: vision, hype and reality for data privacy and protection.
\newblock \emph{IEEE Transactions on Knowledge and Data Engineering}, 2021{\natexlab{b}}.

\bibitem[Liu et~al.(2022{\natexlab{a}})Liu, Kang, Zou, Pu, He, Ye, Ouyang, Zhang, and Yang]{liu2022vertical}
Yang Liu, Yan Kang, Tianyuan Zou, Yanhong Pu, Yuanqin He, Xiaozhou Ye, Ye~Ouyang, Ya-Qin Zhang, and Qiang Yang.
\newblock Vertical federated learning.
\newblock \emph{arXiv preprint arXiv:2211.12814}, 2022{\natexlab{a}}.

\bibitem[Yu et~al.(2024)Yu, Han, Li, Lin, Zhang, Zhang, Liu, Weng, Jeon, Chow, et~al.]{yu2024survey}
Lei Yu, Meng Han, Yiming Li, Changting Lin, Yao Zhang, Mingyang Zhang, Yan Liu, Haiqin Weng, Yuseok Jeon, Ka-Ho Chow, et~al.
\newblock A survey of privacy threats and defense in vertical federated learning: From model life cycle perspective.
\newblock \emph{arXiv preprint arXiv:2402.03688}, 2024.

\bibitem[Yang et~al.(2023)Yang, Chai, Zhang, Jin, Wang, Liu, Tian, Xu, and Chen]{yang2023survey}
Liu Yang, Di~Chai, Junxue Zhang, Yilun Jin, Leye Wang, Hao Liu, Han Tian, Qian Xu, and Kai Chen.
\newblock A survey on vertical federated learning: From a layered perspective.
\newblock \emph{arXiv preprint arXiv:2304.01829}, 2023.

\bibitem[Ye et~al.(2024)Ye, Shen, Du, Snezhko, Kovalev, and Yuen]{ye2024vertical}
Mang Ye, Wei Shen, Bo~Du, Eduard Snezhko, Vassili Kovalev, and Pong~C Yuen.
\newblock Vertical federated learning for effectiveness, security, applicability: A survey.
\newblock \emph{ACM Computing Surveys}, 2024.

\bibitem[Chatel et~al.(2021)Chatel, Pyrgelis, Troncoso-Pastoriza, and Hubaux]{chatel2021sok}
Sylvain Chatel, Apostolos Pyrgelis, Juan~Ram{\'o}n Troncoso-Pastoriza, and Jean-Pierre Hubaux.
\newblock Sok: Privacy-preserving collaborative tree-based model learning.
\newblock \emph{Proceedings on Privacy Enhancing Technologies}, 2021\penalty0 (3):\penalty0 182--203, 2021.

\bibitem[Ong et~al.(2022)Ong, Baracaldo, and Zhou]{ong2022tree}
Yuya~Jeremy Ong, Nathalie Baracaldo, and Yi~Zhou.
\newblock Tree-based models for federated learning systems.
\newblock In \emph{Federated Learning}, pages 27--52. 2022.

\bibitem[Han et~al.(2022)Han, Du, and Yang]{han2022fedgbf}
Yujin Han, Pan Du, and Kai Yang.
\newblock Fedgbf: An efficient vertical federated learning framework via gradient boosting and bagging.
\newblock \emph{arXiv preprint arXiv:2204.00976}, 2022.

\bibitem[Xie et~al.(2022)Xie, Liu, Lu, Chang, and Shi]{xie2022efficient}
Lunchen Xie, Jiaqi Liu, Songtao Lu, Tsung-Hui Chang, and Qingjiang Shi.
\newblock An efficient learning framework for federated xgboost using secret sharing and distributed optimization.
\newblock \emph{ACM Transactions on Intelligent Systems and Technology (TIST)}, 13\penalty0 (5):\penalty0 1--28, 2022.

\bibitem[Wu et~al.(2020)Wu, Cai, Xiao, Chen, and Ooi]{wu2020privacy}
Yuncheng Wu, Shaofeng Cai, Xiaokui Xiao, Gang Chen, and Beng~Chin Ooi.
\newblock Privacy preserving vertical federated learning for tree-based models.
\newblock \emph{Proceedings of the VLDB Endowment}, 13\penalty0 (12):\penalty0 2090--2103, 2020.

\bibitem[Liu et~al.(2022{\natexlab{b}})Liu, Shi, Xie, Li, Hu, Kim, Xu, Li, and Song]{liu2022unifed}
Xiaoyuan Liu, Tianneng Shi, Chulin Xie, Qinbin Li, Kangping Hu, Haoyu Kim, Xiaojun Xu, Bo~Li, and Dawn Song.
\newblock Unifed: A benchmark for federated learning frameworks.
\newblock \emph{arXiv preprint arXiv:2207.10308}, 2022{\natexlab{b}}.

\bibitem[Liu et~al.(2021)Liu, Fan, Chen, Xu, and Yang]{liu2021fate}
Yang Liu, Tao Fan, Tianjian Chen, Qian Xu, and Qiang Yang.
\newblock Fate: An industrial grade platform for collaborative learning with data protection.
\newblock \emph{The Journal of Machine Learning Research}, 22\penalty0 (1):\penalty0 10320--10325, 2021.

\bibitem[Li et~al.(2023)Li, Wu, Cai, Han, Yung, Fu, and He]{fedtree}
Qinbin Li, Zhaomin Wu, Yanzheng Cai, Yuxuan Han, Ching~Man Yung, Tianyuan Fu, and Bingsheng He.
\newblock Fedtree: A federated learning system for trees.
\newblock In \emph{Proceedings of Machine Learning and Systems}, 2023.

\bibitem[Xie et~al.(2023)Xie, Wang, Gao, Chen, Yao, Kuang, Li, Ding, and Zhou]{xie2022federatedscope}
Yuexiang Xie, Zhen Wang, Dawei Gao, Daoyuan Chen, Liuyi Yao, Weirui Kuang, Yaliang Li, Bolin Ding, and Jingren Zhou.
\newblock Federatedscope: A flexible federated learning platform for heterogeneity.
\newblock \emph{Proceedings of the VLDB Endowment}, 16\penalty0 (5):\penalty0 1059--1072, 2023.

\bibitem[Pinkas et~al.(2014)Pinkas, Schneider, and Zohner]{pinkas2014faster}
Benny Pinkas, Thomas Schneider, and Michael Zohner.
\newblock Faster private set intersection based on $\{$OT$\}$ extension.
\newblock In \emph{23rd $\{$USENIX$\}$ Security Symposium ($\{$USENIX$\}$ Security 14)}, pages 797--812, 2014.

\bibitem[Pinkas et~al.(2018)Pinkas, Schneider, and Zohner]{pinkas2018scalable}
Benny Pinkas, Thomas Schneider, and Michael Zohner.
\newblock Scalable private set intersection based on ot extension.
\newblock \emph{ACM Transactions on Privacy and Security (TOPS)}, 21\penalty0 (2):\penalty0 1--35, 2018.

\bibitem[Dong et~al.(2013)Dong, Chen, and Wen]{dong2013private}
Changyu Dong, Liqun Chen, and Zikai Wen.
\newblock When private set intersection meets big data: an efficient and scalable protocol.
\newblock In \emph{Proceedings of the 2013 ACM SIGSAC conference on Computer \& communications security}, pages 789--800, 2013.

\bibitem[Chen et~al.(2017)Chen, Laine, and Rindal]{chen2017fast}
Hao Chen, Kim Laine, and Peter Rindal.
\newblock Fast private set intersection from homomorphic encryption.
\newblock In \emph{Proceedings of the 2017 ACM SIGSAC Conference on Computer and Communications Security}, pages 1243--1255, 2017.

\bibitem[Quinlan(1986)]{quinlan1986induction}
J.~Ross Quinlan.
\newblock Induction of decision trees.
\newblock \emph{Machine learning}, 1:\penalty0 81--106, 1986.

\bibitem[Quinlan(2014)]{quinlan2014c4}
J~Ross Quinlan.
\newblock \emph{C4. 5: programs for machine learning}.
\newblock 2014.

\bibitem[Breiman(2017)]{breiman2017classification}
Leo Breiman.
\newblock \emph{Classification and regression trees}.
\newblock 2017.

\bibitem[Maimon and Rokach(2014)]{maimon2014data}
Oded~Z Maimon and Lior Rokach.
\newblock \emph{Data mining with decision trees: theory and applications}, volume~81.
\newblock 2014.

\bibitem[Song and Ying(2015)]{song2015decision}
Yan-Yan Song and LU~Ying.
\newblock Decision tree methods: applications for classification and prediction.
\newblock \emph{Shanghai archives of psychiatry}, 27\penalty0 (2):\penalty0 130, 2015.

\bibitem[Safavian and Landgrebe(1991)]{safavian1991survey}
S~Rasoul Safavian and David Landgrebe.
\newblock A survey of decision tree classifier methodology.
\newblock \emph{IEEE transactions on systems, man, and cybernetics}, 21\penalty0 (3):\penalty0 660--674, 1991.

\bibitem[Rokach and Maimon(2005)]{rokach2005top}
Lior Rokach and Oded Maimon.
\newblock Top-down induction of decision trees classifiers-a survey.
\newblock \emph{IEEE Transactions on Systems, Man, and Cybernetics, Part C (Applications and Reviews)}, 35\penalty0 (4):\penalty0 476--487, 2005.

\bibitem[Sharma et~al.(2016)Sharma, Kumar, et~al.]{sharma2016survey}
Himani Sharma, Sunil Kumar, et~al.
\newblock A survey on decision tree algorithms of classification in data mining.
\newblock \emph{International Journal of Science and Research (IJSR)}, 5\penalty0 (4):\penalty0 2094--2097, 2016.

\bibitem[Breiman(2001)]{breiman2001random}
Leo Breiman.
\newblock Random forests.
\newblock \emph{Machine learning}, 45\penalty0 (1):\penalty0 5--32, 2001.

\bibitem[Friedman(2001)]{friedman2001greedy}
Jerome~H Friedman.
\newblock Greedy function approximation: a gradient boosting machine.
\newblock \emph{Annals of statistics}, pages 1189--1232, 2001.

\bibitem[Si et~al.(2017)Si, Zhang, Keerthi, Mahajan, Dhillon, and Hsieh]{si2017gradient}
Si~Si, Huan Zhang, S~Sathiya Keerthi, Dhruv Mahajan, Inderjit~S Dhillon, and Cho-Jui Hsieh.
\newblock Gradient boosted decision trees for high dimensional sparse output.
\newblock In \emph{International conference on machine learning}, pages 3182--3190, 2017.

\bibitem[Kasiviswanathan et~al.(2011)Kasiviswanathan, Lee, Nissim, Raskhodnikova, and Smith]{kasiviswanathan2011can}
Shiva~Prasad Kasiviswanathan, Homin~K Lee, Kobbi Nissim, Sofya Raskhodnikova, and Adam Smith.
\newblock What can we learn privately?
\newblock \emph{SIAM Journal on Computing}, 40\penalty0 (3):\penalty0 793--826, 2011.

\bibitem[Yang et~al.(2020)Yang, Lyu, Zhao, Zhu, and Lam]{yang2020local}
Mengmeng Yang, Lingjuan Lyu, Jun Zhao, Tianqing Zhu, and Kwok-Yan Lam.
\newblock Local differential privacy and its applications: A comprehensive survey.
\newblock \emph{arXiv preprint arXiv:2008.03686}, 2020.

\bibitem[Arachchige et~al.(2019)Arachchige, Bertok, Khalil, Liu, Camtepe, and Atiquzzaman]{arachchige2019local}
Pathum Chamikara~Mahawaga Arachchige, Peter Bertok, Ibrahim Khalil, Dongxi Liu, Seyit Camtepe, and Mohammed Atiquzzaman.
\newblock Local differential privacy for deep learning.
\newblock \emph{IEEE Internet of Things Journal}, 7\penalty0 (7):\penalty0 5827--5842, 2019.

\bibitem[Cormode et~al.(2018)Cormode, Jha, Kulkarni, Li, Srivastava, and Wang]{cormode2018privacy}
Graham Cormode, Somesh Jha, Tejas Kulkarni, Ninghui Li, Divesh Srivastava, and Tianhao Wang.
\newblock Privacy at scale: Local differential privacy in practice.
\newblock In \emph{Proceedings of the 2018 International Conference on Management of Data}, pages 1655--1658, 2018.

\bibitem[Alvim et~al.(2018)Alvim, Chatzikokolakis, Palamidessi, and Pazii]{alvim2018local}
M{\'a}rio Alvim, Konstantinos Chatzikokolakis, Catuscia Palamidessi, and Anna Pazii.
\newblock Local differential privacy on metric spaces: optimizing the trade-off with utility.
\newblock In \emph{2018 IEEE 31st Computer Security Foundations Symposium (CSF)}, pages 262--267, 2018.

\bibitem[Chatzikokolakis et~al.(2013)Chatzikokolakis, Andr{\'e}s, Bordenabe, and Palamidessi]{chatzikokolakis2013broadening}
Konstantinos Chatzikokolakis, Miguel~E Andr{\'e}s, Nicol{\'a}s~Emilio Bordenabe, and Catuscia Palamidessi.
\newblock Broadening the scope of differential privacy using metrics.
\newblock In \emph{International Symposium on Privacy Enhancing Technologies Symposium}, pages 82--102, 2013.

\bibitem[He et~al.(2014)He, Machanavajjhala, and Ding]{he2014blowfish}
Xi~He, Ashwin Machanavajjhala, and Bolin Ding.
\newblock Blowfish privacy: Tuning privacy-utility trade-offs using policies.
\newblock In \emph{Proceedings of the 2014 ACM SIGMOD international conference on Management of data}, pages 1447--1458, 2014.

\bibitem[Mohassel and Rindal(2018)]{mohassel2018aby3}
Payman Mohassel and Peter Rindal.
\newblock Aby3: A mixed protocol framework for machine learning.
\newblock In \emph{Proceedings of the 2018 ACM SIGSAC conference on computer and communications security}, pages 35--52, 2018.

\bibitem[Garfinkel et~al.(2003)Garfinkel, Pfaff, Chow, Rosenblum, and Boneh]{garfinkel2003terra}
Tal Garfinkel, Ben Pfaff, Jim Chow, Mendel Rosenblum, and Dan Boneh.
\newblock Terra: A virtual machine-based platform for trusted computing.
\newblock In \emph{Proceedings of the nineteenth ACM symposium on Operating systems principles}, pages 193--206, 2003.

\bibitem[Sabt et~al.(2015)Sabt, Achemlal, and Bouabdallah]{sabt2015trusted}
Mohamed Sabt, Mohammed Achemlal, and Abdelmadjid Bouabdallah.
\newblock Trusted execution environment: what it is, and what it is not.
\newblock In \emph{2015 IEEE Trustcom/BigDataSE/Ispa}, volume~1, pages 57--64, 2015.

\bibitem[Ke et~al.(2017)Ke, Meng, Finley, Wang, Chen, Ma, Ye, and Liu]{ke2017lightgbm}
Guolin Ke, Qi~Meng, Thomas Finley, Taifeng Wang, Wei Chen, Weidong Ma, Qiwei Ye, and Tie-Yan Liu.
\newblock Lightgbm: A highly efficient gradient boosting decision tree.
\newblock \emph{Advances in neural information processing systems}, 30, 2017.

\bibitem[Dorogush et~al.(2018)Dorogush, Ershov, and Gulin]{dorogush2018catboost}
Anna~Veronika Dorogush, Vasily Ershov, and Andrey Gulin.
\newblock Catboost: gradient boosting with categorical features support.
\newblock \emph{arXiv preprint arXiv:1810.11363}, 2018.

\bibitem[Yao et~al.(2022)Yao, Wang, Dai, Bo, and Chen]{yao2022efficient}
Houpu Yao, Jiazhou Wang, Peng Dai, Liefeng Bo, and Yanqing Chen.
\newblock An efficient and robust system for vertically federated random forest.
\newblock \emph{arXiv preprint arXiv:2201.10761}, 2022.

\bibitem[He et~al.(2020{\natexlab{b}})He, Li, So, Zhang, Wang, Wang, Vepakomma, Singh, Qiu, Shen, Zhao, Kang, Liu, Raskar, Yang, Annavaram, and Avestimehr]{fedml}
Chaoyang He, Songze Li, Jinhyun So, Mi~Zhang, Hongyi Wang, Xiaoyang Wang, Praneeth Vepakomma, Abhishek Singh, Hang Qiu, Li~Shen, Peilin Zhao, Yan Kang, Yang Liu, Ramesh Raskar, Qiang Yang, Murali Annavaram, and Salman Avestimehr.
\newblock Fedml: A research library and benchmark for federated machine learning.
\newblock \emph{Advances in Neural Information Processing Systems, Best Paper Award at Federate Learning Workshop}, 2020{\natexlab{b}}.

\bibitem[Lai et~al.(2022)Lai, Dai, Singapuram, Liu, Zhu, Madhyastha, and Chowdhury]{fedscale}
Fan Lai, Yinwei Dai, Sanjay~S. Singapuram, Jiachen Liu, Xiangfeng Zhu, Harsha~V. Madhyastha, and Mosharaf Chowdhury.
\newblock {FedScale}: Benchmarking model and system performance of federated learning at scale.
\newblock In \emph{International Conference on Machine Learning (ICML)}, 2022.

\bibitem[Beutel et~al.(2020)Beutel, Topal, Mathur, Qiu, Fernandez-Marques, Gao, Sani, Kwing, Parcollet, Gusmão, and Lane]{flower}
Daniel~J Beutel, Taner Topal, Akhil Mathur, Xinchi Qiu, Javier Fernandez-Marques, Yan Gao, Lorenzo Sani, Hei~Li Kwing, Titouan Parcollet, Pedro PB~de Gusmão, and Nicholas~D Lane.
\newblock Flower: A friendly federated learning research framework.
\newblock \emph{arXiv preprint arXiv:2007.14390}, 2020.

\bibitem[Wang et~al.(2022)Wang, Kuang, Xie, Yao, Li, Ding, and Zhou]{wang2022federatedscope}
Zhen Wang, Weirui Kuang, Yuexiang Xie, Liuyi Yao, Yaliang Li, Bolin Ding, and Jingren Zhou.
\newblock Federatedscope-gnn: Towards a unified, comprehensive and efficient package for federated graph learning.
\newblock In \emph{Proceedings of the 28th ACM SIGKDD Conference on Knowledge Discovery and Data Mining}, page 4110–4120, 2022.

\bibitem[Kuang et~al.(2024)Kuang, Qian, Li, Chen, Gao, Pan, Xie, Li, Ding, and Zhou]{kuang2024fsllm}
Weirui Kuang, Bingchen Qian, Zitao Li, Daoyuan Chen, Dawei Gao, Xuchen Pan, Yuexiang Xie, Yaliang Li, Bolin Ding, and Jingren Zhou.
\newblock Federatedscope-llm: A comprehensive package for fine-tuning large language models in federated learning.
\newblock In \emph{Proceedings of the 30th ACM SIGKDD Conference on Knowledge Discovery and Data Mining}, page 5260–5271, 2024.

\bibitem[Lim et~al.(2020)Lim, Luong, Hoang, Jiao, Liang, Yang, Niyato, and Miao]{lim2020federated}
Wei Yang~Bryan Lim, Nguyen~Cong Luong, Dinh~Thai Hoang, Yutao Jiao, Ying-Chang Liang, Qiang Yang, Dusit Niyato, and Chunyan Miao.
\newblock Federated learning in mobile edge networks: A comprehensive survey.
\newblock \emph{IEEE Communications Surveys \& Tutorials}, 22\penalty0 (3):\penalty0 2031--2063, 2020.

\bibitem[Hardy et~al.(2017)Hardy, Henecka, Ivey-Law, Nock, Patrini, Smith, and Thorne]{hardy2017private}
Stephen Hardy, Wilko Henecka, Hamish Ivey-Law, Richard Nock, Giorgio Patrini, Guillaume Smith, and Brian Thorne.
\newblock Private federated learning on vertically partitioned data via entity resolution and additively homomorphic encryption.
\newblock \emph{arXiv preprint arXiv:1711.10677}, 2017.

\bibitem[Zhao et~al.(2022)Zhao, Zhu, Xu, Wang, Lu, and Li]{zhao2022sgboost}
Jiaqi Zhao, Hui Zhu, Wei Xu, Fengwei Wang, Rongxing Lu, and Hui Li.
\newblock Sgboost: An efficient and privacy-preserving vertical federated tree boosting framework.
\newblock \emph{IEEE Transactions on Information Forensics and Security}, 18:\penalty0 1022--1036, 2022.

\end{thebibliography}

\end{document}